\documentclass[acmtog, nonacm]{acmart}
\setcopyright{none}
\renewcommand\footnotetextcopyrightpermission[1]{}

\usepackage{booktabs} 

\definecolor{amber}{rgb}{1.0, 0.3, 0}
\definecolor{green}{rgb}{0, 0.7, 0}
\newcommand{\ccd}[1]{{\color{black}{#1}}}
\newcommand{\zy}[1]{{\color{black}{#1}}}

\newcommand{\cd}[1]{{\color{black}{#1}}}

\newcommand{\eg}[1]{{\textit{e.g.,~}}}
\newcommand{\ie}[1]{{\textit{i.e.,~}}}

\usepackage{booktabs}
\usepackage[ruled]{algorithm2e} 
\usepackage{graphicx}
\usepackage{float}
\usepackage{stfloats}
\usepackage{amsmath}
\usepackage{makecell}
\usepackage{comment}
\usepackage{subcaption}
\usepackage{hyperref}
\usepackage{caption}

\usepackage[ruled]{algorithm2e} 

\SetAlFnt{\small}
\SetAlCapFnt{\small}
\SetAlCapNameFnt{\small}
\SetAlCapHSkip{0pt}

\acmJournal{TOG}

\begin{document}
\title{SGEdit: Bridging LLM with Text2Image Generative Model for Scene Graph-based Image Editing}

\author{Zhiyuan Zhang}
\email{zzhang452-c@my.cityu.edu.hk}
\affiliation{%
 \institution{City University of Hong Kong}
 \city{Hong Kong SAR}
 \country{China}}

\author{DongDong Chen}
\email{cddlyf@gmail.com}
\affiliation{%
 \institution{Microsoft GenAI}
 \city{Redmond}
 \country{US}}
 
\author{Jing Liao}
\email{jingliao@cityu.edu.hk}
\affiliation{
 \institution{City University of Hong Kong}
 \city{Hong Kong SAR}
 \country{China}}
\authornote{Corresponding author.}
 


\newcommand{\ddc}[1]{\textcolor{black}{[DDC: #1]}}


\begin{abstract}
\par

Scene graphs offer a structured, hierarchical representation of images, with nodes and edges symbolizing objects and the relationships among them. It can serve as a natural interface for image editing, dramatically improving precision and flexibility. Leveraging this benefit, we introduce a new framework that integrates large language model (LLM) with Text2Image generative model for scene graph-based image editing. This integration enables precise modifications at the object level and creative recomposition of scenes without compromising overall image integrity. Our approach involves two primary stages: 1) Utilizing a LLM-driven scene parser, we construct an image's scene graph, capturing key objects and their interrelationships, as well as parsing fine-grained attributes such as object masks and descriptions. These annotations facilitate concept learning with a fine-tuned diffusion model, representing each object with an optimized token and detailed description prompt. 2) During the image editing phase, a LLM editing controller guides the edits towards specific areas. These edits are then implemented by an attention-modulated diffusion editor, utilizing the fine-tuned model to perform object additions, deletions, replacements, and adjustments. Through extensive experiments, we demonstrate that our framework significantly outperforms existing image editing methods in terms of editing precision and scene aesthetics. Our code is available at https://bestzzhang.github.io/SGEdit.
\end{abstract}



\begin{teaserfigure}
    \centering
    \includegraphics[width=1.0\textwidth]{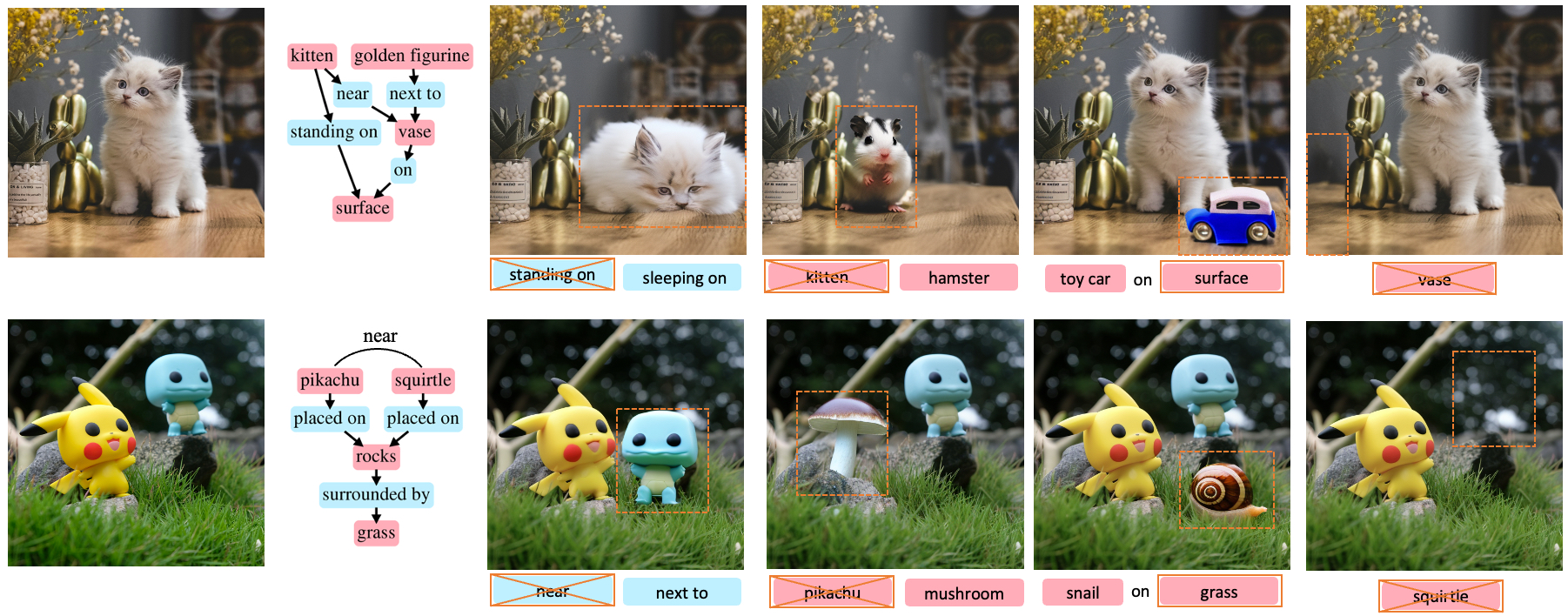}
    \caption{Our approach predicts a scene graph as a user interface, enabling modifications to nodes and edges for various tasks, such as changing relationships or replacing, adding, and removing elements. By integrating LLM and Text2Image generative models, users can explore diverse compositions in their images, ensuring that these alterations accurately reflect the structural changes depicted in the modified scene graph. Input images are sourced from \textcopyright Unsplash.}
    \label{fig:teaser}
\end{teaserfigure}
\maketitle

\section{Introduction}

Scene graph is a data structure commonly used in computer graphics and image processing to represent the elements of a scene in a hierarchical manner~\cite{krishna2017visual, johnson2015image}. It organizes objects and their relationships in a graph-like structure, where nodes represent objects, and edges define the relationships among these objects. Therefore, utilizing scene graphs as image editing interface offers multiple benefits. First, it facilitates high-level modifications, enabling precise edits on specific elements without impacting the entire image. Second, it allows users to explore creative compositions or scenarios through straightforward alterations to the scene graph, bypassing the need for intricate mask creation or extensive prompt formulation. Third, it delivers a cohesive interface for various editing tasks, including the addition, removal, or substitution of objects, as well as altering their relationships.

 Existing scene graph-based image editing methods often employ a scene graph predictor ~\cite{dhamo2020semantic} to construct the scene-graph or directly experiment on images with scene graph annotations~\cite{su2021fully, zhang2023complex}, then train a layout predictor based on the modified scene graph, which is finally used as the guidance to generate the edited result. The early exploration is mainly Generative Adversarial Networks (GANs) based ~\cite{dhamo2020semantic, su2021fully}. However, the image quality generated by these GAN-based methods is generally inferior, and the range of editable objects is confined to those within the close-set categories supported by the scene-graph predictor. Recently, large-scale Text2Image diffusion models ~\cite{rombach_high-resolution_2022,zhao2023uni,fan2023frido} have demonstrated better generation quality and have enabled many applications in image editing~\cite{hertz_prompt--prompt_2022, gal_image_2022, brooks_instructpix2pix_2023, kawar_imagic_2023, yang2023paint, cao2023masactrl}. One diffusion-based approach\cd{~\cite{zhang2023complex}}, explored in parallel with our research, adopted a similar training strategy as the above GANs based methods~\cite{dhamo2020semantic}. Despite the improved image quality, it still suffers from limited category support and cannot well handle in-the-wild images. In this paper, we propose a new scene graph-based editing framework that innovatively integrates LLM and a Text2Image Generative model, which can avoid the above limitation and support editing across various categories. The LLM\footnote{In this paper, our method uses LLM as the general name for Large Language Models and Multimodal Large Language Models}~\cite{OpenAI_2023} serves not only as the general scene parser but also as the intelligent editing controller, while the Text2Image Diffusion model~\cite{rombach_high-resolution_2022} is customized to perform high-quality, region-wise image editing.
 

More specifically, our framework consists of two main stages: scene parsing and image editing. In the scene parsing stage, the user-uploaded image is processed by our LLM-driven scene parser, which constructs a scene graph and annotates each node (i.e., object) with fine-grained attributes including the object mask and detailed caption. Based on these node annotations, we subsequently learn the multiple concepts in the image by fine-tuning a diffusion model, representing each object not only with a distinct token but also with a specific, detailed caption prompt. This key new design differentiate us from existing methods that employ either a single text token ~\cite{ruiz_dreambooth_2023, gal_image_2022} or a detailed description ~\cite{wei2023diffusion, yang2023idea2img}. As a detailed description can encapsulate the general semantics/attributes of an object, it simplifies token optimization by focusing on the residual. This, in turn, eases the learning process, better preserves the visual properties, and improves editability.


In the image editing stage, users can interact with the constructed scene graph for various operations, such as changing relationships, adding or removing objects in the image. We formulate all operations as sequential "remove/generate" steps, controlled by an LLM editing controller and executed by our attention-modulated diffusion editor. Inspired by recent works employing LLMs for planning in visual tasks\cite{lian2023llmImg, qu2023layoutllm, feng2024layoutgpt, lian2023llm, lu2023flowzero}, we utilize the LLM as the editing controller. The LLM not only analyzes the scene graph and user modifications to identify the objects to be removed and created, but also suggests the target generation area along with the corresponding text prompts. To achieve the desired modification, we utilize the masks for the objects to be removed and the suggested bounding boxes for the generated objects for object removal and insertion. By modulating the attention in the sampling process, we query the content in the preserved region to fill in the erased region for seamless object removal. We also enhance the spatial alignment between the text and target region to guide object insertion. By integrating the LLM with the diffusion model, we facilitate accurate and context-aware image edits.

We conduct comprehensive experiments and perceptual studies to demonstrate the superiority of our proposed method over existing baseline methods. Due to the absence of standardized automatic metrics for this task, we propose the use of GPT-4V as an evaluator to assess composition elements, alignment of relations, and overall image quality. Additionally, we corroborate our findings through human evaluations. We also introduce a user interface designed for interactive scene graph editing, enhancing the design process.

To summarize, our contributions are as follows:
\begin{itemize}
    \item Introducing a scene graph-based image manipulation framework that integrates LLM and Text2Image generative model, enabling editing across a broad range of descriptive terms. And an interactive user interface is provided for ease of use.
    \item Utilizing LLM for effective scene parsing and editing control, ensuring flexible and context-aware image edits.
    \item Using the diffusion model for learning multiple concepts with both detailed description prompts and optimized tokens. And performing various scene graph operations uniformly with Attention-modulated Object Removal and Insertion.
\end{itemize}

\section{Related Works}
\subsection{Scene Graph to Image Synthesis and Editing}
Scene graphs are diagrams where objects in a scene are represented as nodes, and the relationships among these objects are shown as edges~\cite{johnson2015image, krishna2017visual}. They offer a powerful structured representation widely used for vision tasks, such as image retrieval~\cite{johnson2015image} and evaluating image captioning~\cite{anderson2016spice}. Johnson et al. \cd{[\citeyear{johnson2018image}]} were the first to adopt it for image synthesis and proposed a dual-phase generation process, which maps the scene graphs to a scene layout and then refines it into a realistic image using a generative model. Following this, various methods have been proposed to generate images from scene graphs, including recent works based on diffusion models~\cite{yang2022diffusion, farshad2023scenegenie, liu2024r3cd}.

Moving beyond pure image generation, several works also employ the scene graph for image editing. The pioneering work in this field, SIMSG~\cite{dhamo2020semantic}, introduced a dual-phase strategy involving layout generation from scene graphs followed by image creation from the generated layout. Initially, a spatial-semantic scene graph network processes the altered scene graph to produce a layout based on bounding boxes. Subsequently, this layout, combined with image features, is input into an image decoder like SPADE for semantic image generation. FFIMSG~\cite{su2021fully}, another GAN-based method, predicts the mask for a specific instance from a subset of the scene graph tied to user modifications and integrates the instance within the specified mask region. In parallel with our research, a diffusion-based method for scene graph based image editing, CSIE~\cite{zhang2023complex}, employs a Region of Interest predictor to determine the bounding box for the region to be modified and uses a diffusion model to inpaint that area. However, all these methods have limited category support, restricting their usage in real-world open-vocabulary applications. 
Different from these approaches, our method integrates LLM and a diffusion-based Text2Image generative model, both of which are a general model, eliminating the limitations on the categories of objects and relationships that can be edited.

\cd{\subsection{LLM-based Image Synthesis}}
Text2Image synthesis benefits from integrating LLMs. The vast world knowledge embedded in these large models enables extended capabilities to create or edit images that are faithful to the user's intention. For example, visual ChatGPT~\cite{wu2023visual} chains ChatGPT with different vision foundations, such as BLIP~\cite{li2022blip}, Stable Diffusion~\cite{rombach_high-resolution_2022}, and ControlNet~\cite{zhang_adding_2023}, enabling multi-step image editing and generation. Idea2Img~\cite{yang2023idea2img} performs self-refinement with GPT-4V (vision) through iterative image selection and prompt adjustment, creating preferable images for users.

A broad range of work is also exploring the use of intermediate representations from external large models to create reasonable spatial arrangements that closely match the given textual prompts. LMD~\cite{lian2023llmImg} leverages the LLM-generated layout to guide image generation in a training-free manner. LayoutGPT~\cite{feng2024layoutgpt} proposes prompting an LLM for layout planning across various domains. \cd{Similarly, Attention Refocusing~\cite{phung2024grounded} uses an LLM for layout generation, improving the alignment with text prompts involving multiple objects or spatial compositions.} RPG~\cite{yang2024mastering} uses \cd{LLM} to convert an input prompt into enriched regional prompts with a planned area, then the generated image is analyzed and refined through additional rounds. \cd{SLD~\cite{wu2024self} proposes using an LLM to parse the input prompt and plan the modification to rectify the misalignment between the generated image and the user prompt.} In light of these findings, this paper proposes the first attempt at utilizing LLM and Text2Image generative model for scene graph based image editing. 


\subsection{Image Editing with Diffusion Models}
By leveraging the power of pretrained Text2Image diffusion models, many image editing methods~\cite{song2020denoising,mokady2023null,cao2023masactrl} have been derived. Many text-driven image editing works~\cite{tumanyan2023plug, hertz_prompt--prompt_2022, cao2023masactrl} directly invert the image into latent noise using methods like DDIM~\cite{song2020denoising} and Null-text inversion~\cite{mokady2023null}, and perform editing by utilizing intermediate features. And Prompt-to-Prompt~\cite{hertz_prompt--prompt_2022} approach proposes to inject the cross-attention maps of the source prompt during target image synthesis to perform both local and global editing. In contrast, MasaCtrl~\cite{cao2023masactrl} proposes the mutual self-attention which replaces the Key and Value matrices of the self-attention in the source image to transfer the visual characteristics of the existing image, enabling non-rigid changes to the object. \cd{Some methods utilize loss derived from the self-attention map to guide image editing. For instance, Diffusion Self-Guidance~\cite{epstein2023diffusion} introduces an attention-based loss that controls various object properties, such as shape, location, and appearance.} There also exist some works that fine-tunes the text embedding~\cite{gal_image_2022} or the U-Net~\cite{ruiz_dreambooth_2023} of the diffusion model to learn concepts within the image, thereby enabling the composition of new scenes with learned concepts by directly changing the prompt. Apart from editing the image using the target text description, InstructPix2Pix~\cite{brooks_instructpix2pix_2023} explores modifying the image with editing instructions by retraining the diffusion model on the \cd{automatically synthesized paired images}.



However, since real-world scenes typically contain multiple objects with very diverse relationships, it is challenging for the above editing methods to precisely identify the region to edit without explicit spatial directives from users. In this paper, we propose to build the explicit scene graph as the editing interface, and leverage the LLM's strong understanding and reasoning capability to automatically determine regions of interest for modification, significantly reducing the users' burden in manually specifying editing regions.  

\section{Method}

\begin{figure*}[t]
    \centering
    \includegraphics[width=1.0\textwidth]{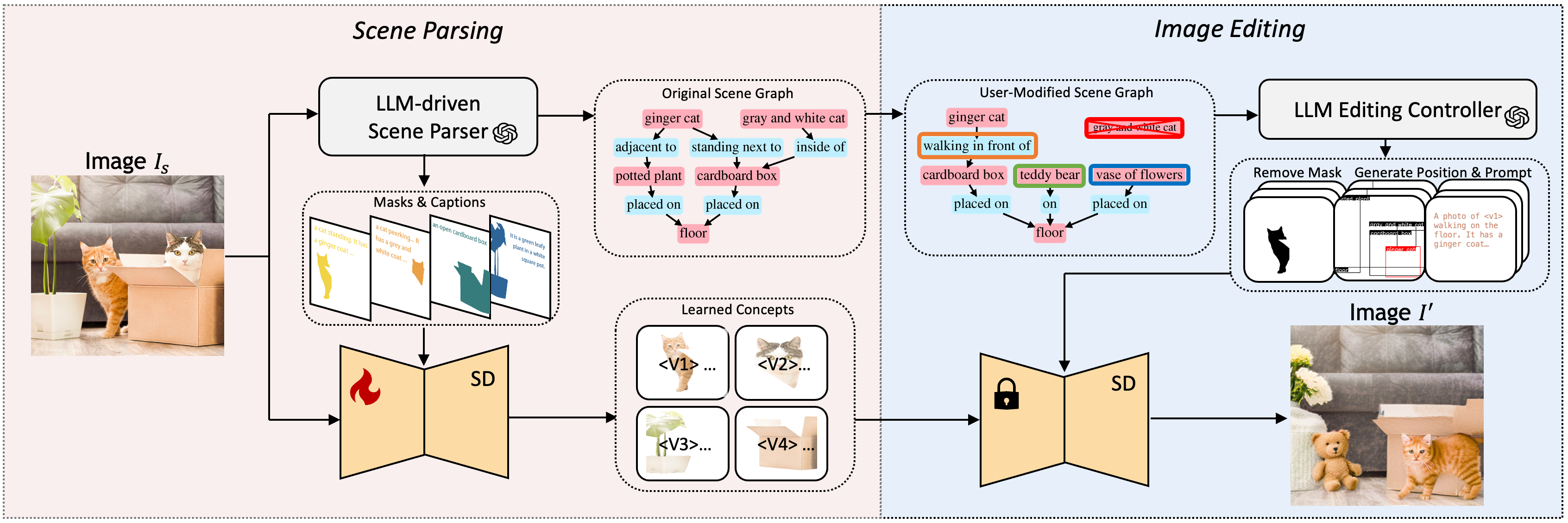}
    \caption{Our pipeline consists of two main stages: scene parsing and image editing. In the scene parsing stage, the input image is processed by our LLM-driven scene parser, which creates a scene graph and annotations for nodes such as object masks and captions. The node annotations allow for the fine-tuning of the diffusion model, representing each object in the scene with an optimized token and a specific prompt. During the image editing stage, the LLM editing controller translates user manipulations on the scene graph into a sequence of operations with text prompts and directs the targeted edits to specific regions. These edits are implemented by applying attention modulation to the fine-tuned diffusion model, enabling object additions, removals, replacements, and relationship modifications in the scene. The input image is from \textcopyright iStockphoto.}
    \label{fig:overall}
\end{figure*}

\subsection{Overview}
\begin{figure}[htbp]
	\centerline{\includegraphics[width=1.0\linewidth]{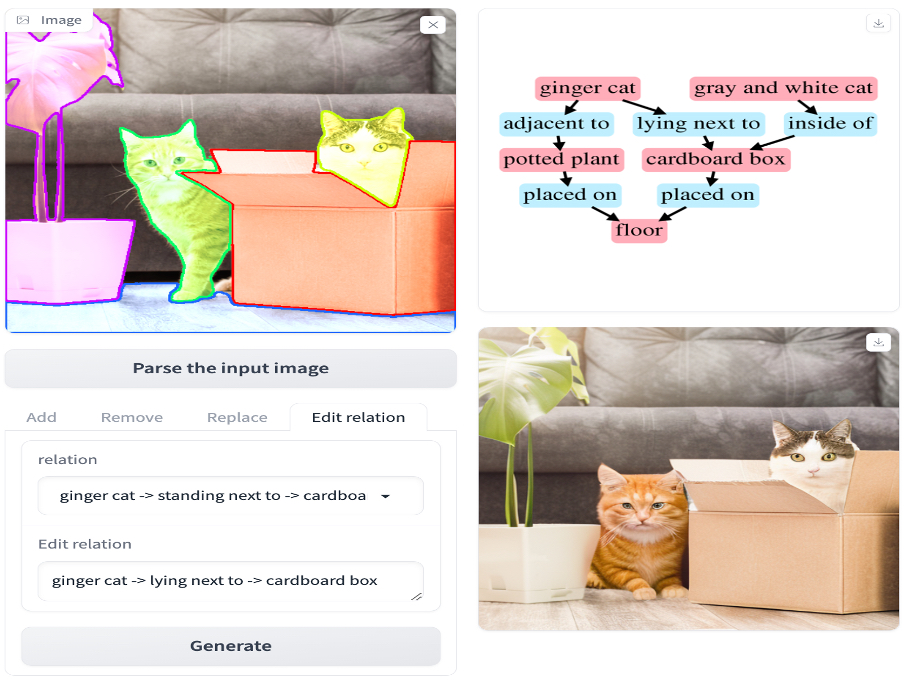}}
	\caption{Screenshot of our interface. The input image is from \textcopyright\ iStockphoto.}
\label{fig:interface}
\end{figure}

Our framework consists of two stages: scene parsing and image editing, as shown in Fig.~\ref{fig:overall}. In the first stage, given the user input image, our LLM-driven scene parser constructs the scene graph and obtains segmentation masks and detailed captions for each object. These masks and captions are used to fine-tune a stable diffusion model, customizing it for the scene and learning the concept of each object with an optimized token and a specific detailed prompt. In the image editing stage, users can modify nodes and edges on the scene graph. The LLM-driven controller targets specific areas for modification and translates user actions on the scene graph into text prompts for a sequence of object removal/insertion operations. Subsequently, we execute the "remove/generate" sequence with our customized diffusion model through attention modulation. Based on our proposed pipeline, we have also built an interactive image editing system with scene graph. Fig. \ref{fig:interface} shows a screenshot of the user interface. Users can perform various editing operations using the bottom-left panel and see the edited result in the right column.

\subsection{Scene Parsing}  \label{sec:preprocess}
Before performing various kinds of editing, our framework is initialized by parsing the input image into a structured representation using a LLM-driven scene parser and learning the concepts of multiple objects from the image with detailed prompts and optimized tokens.

\subsubsection{LLM-driven Scene Parser}
Our LLM-driven Scene Parser translates visual inputs into a structured scene graph, accompanied by detailed node attribute annotations. This initial transformation sets the groundwork for subsequent concept learning and manipulation tasks. \cd{To achieve this, we decompose the scene parsing task into a series of targeted questions, extracting the necessary scene information through multiple rounds of interaction with the LLM automatically. We design carefully crafted prompt templates with in-context examples to guide the LLM's responses, ensuring it outputs the desired information in a consistent manner. }


\textbf{Scene Graph Construction: } The scene parsing process begins by constructing a scene graph $\mathcal{SG}(O, R)$ from an input image $I_s$. Here, nodes \(O\) encapsulates \({n_O}\) detected objects \(\{o_1, o_2, \ldots, o_{n_O}\}\), while edges \(R\) represents a total of \({n_R}\) relationships \(\{r_1, r_2, \ldots, r_{n_R}\}\) among these objects. Each relationship is expressed as a triple \((o_i, r_k, o_j)\), denoting the subject, predicate, and object, respectively. Initially, the LLM is directly commanded to describe the scene, which is followed by more focused tasks guided by specifically crafted in-context examples. These tasks include identifying “what are the main instances” and “what are the relationships between these instances,” which are used to build the scene graph. The LLM is also guided to adhere to specific requirements, such as specifying foreground instances (e.g., differentiating between a "ginger cat" and a "gray and white cat" as shown in Fig.~\ref{fig:overall}), excluding accessory parts (e.g., "saddles" on "horse"), and simplifying the background into basic elements (e.g., "field", "floor").

\textbf{Node Attribute Extraction: } Subsequent to the construction of the scene graph, the scene parser further extracts crucial attributes for each node $o_i$, including its segmentation mask \(m_i \in M\), bounding box \(b_i \in B\), and detailed descriptions \(t_i \in T\). We employ an open-vocabulary segmenter, Grounded-SAM ~\cite{ren2024grounded}, to delineate the spatial boundaries of the object nodes in the scene graph. The LLM is then queried again to generate a detailed description for each detected object. These descriptions will be formatted and utilized in following concept learning (Section~\ref{sec:fine-tune}).

\subsubsection{Concept Learning with Both Detailed Prompt and Textual Inversion}  \label{sec:fine-tune}

One key requirement for image editing is to preserve the original visual features of objects in the input image. This can be achieved by learning the concept of each object in the Text2Image diffusion model. To this end, recent works have proposed two main approaches: representing an object with a single text token through textual inversion~\cite{ruiz_dreambooth_2023, gal_image_2022}, or directly using information-rich text without textual inversion~\cite{wei2023diffusion, yang2023idea2img}. The first approach involves optimizing text embeddings and fine-tuning the model, which may lead to overfitting to irrelevant details, such as object pose, thus constraining editability. The second approach transforms an object into text that is rich in detailed semantic concepts, offering more flexibility in editing via text but cannot perfectly reconstruct the original object, because text is not enough to convey some fine-grained appearance details.

\begin{figure}[h]
\centerline{\includegraphics[width=1.0\linewidth]{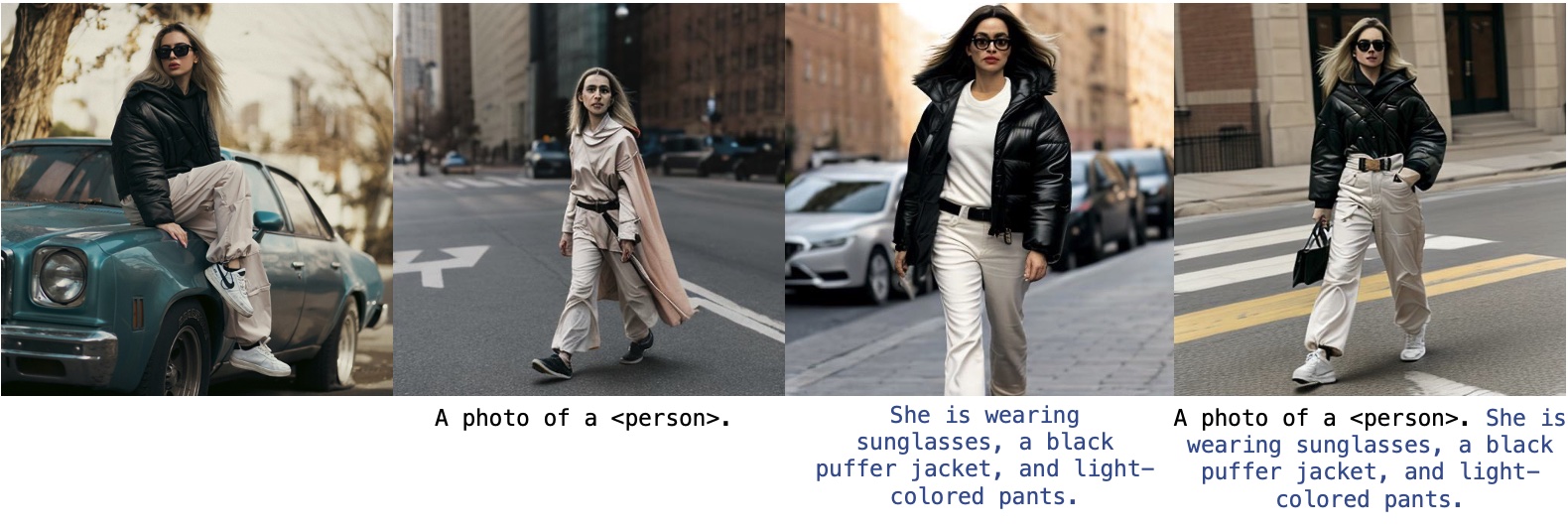}}
	\caption{Contents represented by the detailed description and optimized token. The leftmost column shows the input images, while the three right columns display images generated using the optimized token, detailed description, and their combination. The combination best preserves the woman's visual identity. The input image is from \textcopyright\ Unsplash.}
	\label{fig:hypothesize}
\end{figure}

Different from existing methods, we propose to leverage the strengths of both the aforementioned approaches by representing each object using detailed prompts along with optimized tokens. In particular, we design a prompt template that comprises two segments: an added token "<opt>" for the subject and a detailed description. The detailed description provides information about the object's general appearance, including its attributes and any accessories attached to it, such as clothing. It can also capture the object's poses, such as standing, running, or performing weightlifting, particularly if the object is an animated \cd{being, while} tokens to be optimized encode residual information that is not  explicitly mentioned in the description.

Following Break-A-Scene~\cite{avrahami2023break}, to optimize the token, we first freeze the model and update the text embedding for the added token. Then we unfreeze the model and optimize the text tokens and the model together. Instead of using "a photo of <opt>." as the training prompt, we adopt the prompt "a photo of <opt>. <detail\_des>", with both the added token "<opt>" and a detailed description "<detail\_des>".

This method has three advantages: Firstly, the detailed description enables residual learning in the tokens optimization, making the token learning process easier. Secondly, by explicitly including the object's appearance in the text, it provides greater consistency in retaining the visual attributes compared to a single optimized text token. Thirdly, decoupling attributes (e.g. pose) from the optimized token helps prevent overfitting and ensures the model's editability. These advantages can be reflected in the example shown in Fig. \ref{fig:hypothesize}. The appearance description "She is wearing sunglasses, a black puffer jacket, and light-colored pants." retains the information of the wearable for the woman, while the optimized token "<opt>" encodes information related to the identity of the woman, such as hair color that is not described. These demonstrates that the detailed description captures specific details pertinent to the text's semantics, while the optimized token aids the model in learning additional residual features. Combining this two part produces the image with the best visual property preservation.

\subsection{Image Editing
}~\label{editing}
With the scene graph representation of the input image, users can conveniently convey their edits by modifying the nodes and edges in the scene graph. Next, our LLM editing controller translates the user's modifications on the scene graph into a sequence of edit operations and directs the targeted edits to specific regions. Then, we utilize the learned concepts and the fine-tuned diffusion model to uniformly perform various types of editing through attention-modulated object removal and insertion approaches.

\subsubsection{LLM Editing Controller} \label{sec:planning}
We propose an editing controller based on LLM to translate user modifications on the scene graph into a sequence of object removal and insertion operations. Additionally, our controller generates prompts based on the scene graph to guide the editing process. \cd{Similar to the LLM-driven scene parser, we adopt prompting templates and example-based learning.}

\textbf{Determine operation sequence:} 
We adopt a simple operation sequence for modifying the scene graph: a removal operation followed by an insertion operation. For example, to change the position of a ginger cat from standing next to a box to walking in front of it (as shown in Fig. ~\ref{fig:overall}), the operation involves "removing" the ginger cat in the input image and then "inserting" it in front of the box with a walking pose. We query the LLM controller to obtain the operation sequence. It processes the original scene graph \(\mathcal{SG}(O, R)\) with bounding boxes \(B\) for each object and the user’s modifications. This yields a set of objects to be removed \(O^{\mathrm{rm}}=\{ o_i^{\mathrm{rm}} \}_{i=1}^{n^{\mathrm{rm}}}\) and a set of objects to be generated \(O^{\mathrm{gen}}=\{ o_i^{\mathrm{gen}} \}_{i=1}^{n^{\mathrm{gen}}}\). For each object $o_i^{\mathrm{rm}}$ to be removed, we obtain its masks \(m_i^{\mathrm{rm}} \) from the scene parsing results. For each object $o_i^{\mathrm{gen}}$ to be generated, the LLM also provides the bounding boxes $b_i^{\mathrm{gen}}$ to specify its location.

\textbf{Generate Text Prompts:} 
We use the same non-object prompt \(y_{non}\)=\textit{`A photo with no objects or people, only the background.'} for all object removal operations. Thus, we only need the controller to generate a text prompt \(y_{gen}\) for each insertion operation. When the objects to be generated contain multiple items, we instruct the LLM to construct a prompt that integrates the objects based on their relationships in the modified scene graph. If a single object is to be generated, the controller crafts a prompt like: "A photo of [generated object] [action] [with support surface]".

\subsubsection{Attention-modulated Object Removal and Insertion} \label{sec:sampler}
Our object removal and insertion operations are achieved through attention modulation with the fine-tuned diffusion model. For better understanding, we will provide a brief review of the attention layers in the diffusion model. Specifically, during the denoising phase at step \( t \), deep image features \( f^l_t \) from convolutional layers are first processed through a self-attention mechanism, which play a crucial role in synthesizing a globally coherent structure. These features are then integrated with textual cues from the given prompt \( P \) via a cross-attention layer. The attention mechanism can be formulated as follows:
\begin{equation}
\text{Attention}(Q, K, V) = \text{softmax}\left(\frac{QK^T}{\sqrt{d}}\right)V
\end{equation}
Here, \(Q\) denotes the query features derived from visual features, while \(K\) and \(V\) represent key and value features, originating either from visual features in self-attention layers or from contextual text embeddings in cross-attention layers. The term \(d\) signifies the dimensionality of these features. 

\begin{figure*}[t]
    \centering
    \includegraphics[width=1.0\textwidth]{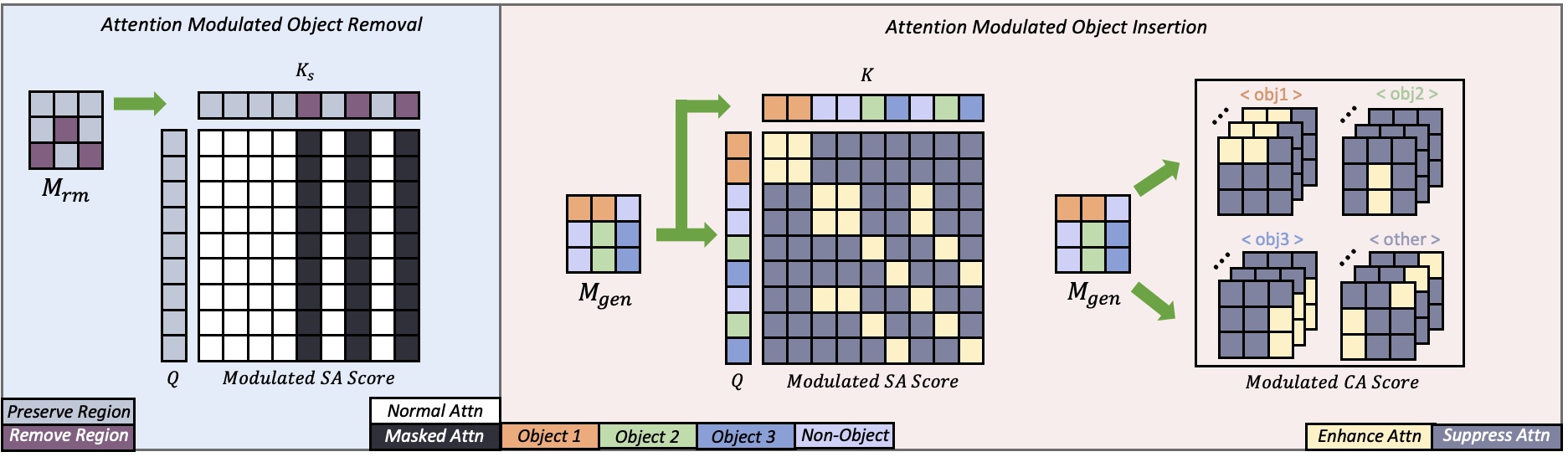}
    \caption{The illustration of our Attention Modulated Object Removal and Insertion. The left part shows the attention modulation in self-attention for object removal, and the right part shows the attention modulation in both self and cross-attention for object insertion.}
    \label{fig:attention}
\end{figure*}

\textbf{Object Removal:}\label{sec:removal}
Our goal is to create a modified version of the input image \(I_s\), denoted as \(I'\), by seamlessly removing the objects defined in \(O^{\mathrm{rm}}\) from the union of masked areas \({M}_{\mathrm{rm}} = \bigcup_{i=1}^{n^{\mathrm{rm}}} m_i^{\mathrm{rm}}\). However, directly generating \(I'\) with the non-object prompt does not ensure that filled region is consistently blended with the unmasked region, without introducing undesired content. Drawing inspiration from mutual self-attention techniques~\cite{cao2023masactrl}, we introduce a modulated attention strategy to query the features from the unmasked region of \(I_s\) to guide the generation of the masked region of \(I'\).


This process begins with a DDIM inversion of the image \(I_s\) to obtain a series of inverted latent states \([z_T^s, \ldots, z_0^s]\). At each step \(t\) of the inversion process, we also extract the Key and Value features \(K_s\) and \(V_s\) in the self-attention layer from the denoised inverted latent \(z_t^s\). We then replace the erased region \({M}_{\mathrm{rm}}\) in \(z_T^s\) with random noise to form the initial latent \(z_T\), which is then fed into the forward sampling process. During the sampling, corresponding \({K, V}\) features of \(z_t\) in the self-attention layer are replaced by two features \(K_s\) and \(V_s\) respectively and the self-attention is modulated as follows:


\begin{gather}
A' = \text{softmax}\left(\frac{Q{K_s}^T + X}{\sqrt{d}}\right) {V_s} \\
X_{ij} = \begin{cases} 
{-\infty}, & {M}_{\mathrm{rm}}[j] = 1 \\
0, & \text{otherwise}
\end{cases}.
\end{gather}

Here we use \(X\) to modulate the self-attention according to the mask \({M}_{\mathrm{rm}}\). As illustrated in the left part of Fig.~\ref{fig:attention}, when \({M}_{\mathrm{rm}}[j]\) is 1, it indicates that the \(j\)-th position in the key is a part to be erased and will not participate in the attention by setting attention with $K_s[j]$ to $0$. Therefore, only the unmasked features will attend to the query features and guide the content generation to be consistent with the unmasked region while erasing the masked region.

\textbf{Object Insertion: }
The goal of object insertion is to generate an image $I'$ with each object $o_i^{\mathrm{gen}}$ within the bounding box $b_i^{\mathrm{gen}}$ predicted by the LLM editing controller, while keeping other parts intact as in $I_s$.

To control the layout of the objects, we modulate the cross-attention to enhance the spatial alignment between the text segments and the corresponding bounding boxes. We also modulate the self-attention to interrupt communication between different bounding boxes and, as a result, restrict the existence of objects to their specified regions only. Following DenseDiffusion \cite{kim2023dense}, the attention mechanism can be formulated as follows:

\begin{gather}
A' = \text{softmax}\left(\frac{QK^T + X}{\sqrt{d}}\right)V \\
X = \lambda_t \cdot \mathcal{R} \odot X_{\text{pos}} \odot (1 - S) \\
- \lambda_t \cdot (1 - \mathcal{R}) \odot X_{\text{neg}} \odot (1 - S)
\end{gather}

Here, \(X\) represents a modulation matrix that adjusts the attention scores based on specific conditions. The matrix \(\mathcal{R} \in \mathbb{R}^{\lvert Q \rvert \times \lvert K \rvert}\) indicates whether the attention scores between query-key pairs should be increased or decreased. The matrices \(X_{\text{pos}}\) and \(X_{\text{neg}}\) are adaptive to positive and negative value ranges respectively, \(S\) adjusts to the size of mask area, and \(\lambda_t\) is a scaling factor that varies with the timestep. 

An intuitive illustration is shown on the right part of Fig.~\ref{fig:attention}. We first generate a mask \(M_{\text{gen}}\) by unionizing all bounding boxes of the inserted objects. Each distinct color in \(M_{\text{gen}}\) corresponds to a semantic segment, including objects and non-objects. In self-attention modulation, the interaction of tokens within the same semantic segment is enhanced, while interaction across different semantic segments is suppressed. In cross-attention modulation, for the object \(o_i^{\mathrm{gen}}\), we enhance the attention value for the corresponding object words within the area \(b_i^{\mathrm{gen}} \) and suppress the attention value outside. For tokens not corresponding to objects, we treat them in the same way in the region corresponding to the complement of the union of all the objects' regions, denoted as \(M_{non}\).

Although the attention modulation control can position the generated objects, we have observed that the objects often extend beyond their bounding boxes. To address this, we generate the noise in \(M_{non}\) region with the non-object prompt \(y_{non}\), and then blend it with the noise guided by the prompt \(y_{gen}\) generated in object regions \(1-M_{non}\), at each sampling step $t$. This is a key difference from DenseDiffusion \cite{kim2023dense}.



Besides generating objects in $I'$, another requirement is to make other regions of $I'$ the same as the input image $I_s$. To achieve it, we divide the sampling process into two phases. In the first phase, during the early steps from \(T\) to \(T_n\), we generate the objects in \(O^{\mathrm{gen}} \) in the positions specified by the bounding boxes \(B^{\mathrm{gen}}\). If there are multiple objects to be generated, we have adopted a multi-instance sampler strategy~\cite{wang2024instancediffusion} to minimize attribute leakage across multiple instances. Specifically, from \(T\) to \(T_m\) (\(T_m > T_n\)), we perform cross-attention modulation for each region using a prompt containing a single object, such as "a photo of [generated object]." instead of \(y_{gen}\) containing multiple object used in \(T_m \) to \(T_n\).  At timestep \(T_n\), the positions of the objects in \(z_n\) have been determined. Then we turn the the second phase. We obtain a precise mask \(M_{seg}\) by predicting latents \(z_0\) from \(z_n\) and use the Grounded-SAM ~\cite{ren2024grounded} to segment the object using bounding boxes on the image decoded from \(z_0\). Then, in the remaining timesteps, we continue to denoise the image while blending it with the existing background of \(I'\) using \(M_{seg}\). The technique allows for a seamless transition between newly inserted objects and untouched backgrounds.

\subsection{Implementation Details}


In our experiment, we used GPT-4V and GPT-4 as the language models (LLM) in the scene parser and editing controller, respectively. We designed task-aware prompting templates and included high-quality examples to facilitate in-context learning. For more information on the detailed prompting design, please refer to the supplemental materials.

We leverage Stable Diffusion v2-1-base in both concept learning and attention-modulated image editing. For concept learning, we dynamically adjust the number of training steps based on the number of objects to be learned. We start with a minimum of 800 steps (for 2 or fewer objects) and increase by 200 steps for each additional object, with a maximum of 1200 steps (for 4 or more objects). \cd{During training, we use the Adam optimizer with learning rates of 5e-4 for updating the text embedding of the added tokens and 2e-6 for optimizing the text tokens and model together. On a single A6000 GPU, this process takes approximately 6 to 10 minutes to complete.} 

In the editing stage, for object removal, we perform attention modulation throughout the entire sampling process. For object insertion, we pre-generate the object with attention modulation from \(t=1.0\) to \(T_n=0.6\) and blend it with the existing background afterwards. If there is more than one object in the generated sets, we adopt a multi-sampler from \(t=1.0\) to \(T_m=0.8\). \cd{The object removal sampling takes 6 seconds, and the object generation sampling takes 5 seconds. The object removal and generation operations are combined to perform high-level user operations, such as relationship modification, with the longest time to perform an operation being less than 15 seconds. Time to access chatGPT is excluded due to the uncertainty affected by network latency.}

\section{Experiments}

\subsection{Experiment Setup} \label{sec:setup}

\begin{figure*}[t]
    \centering
    \includegraphics[width=1.0\textwidth]{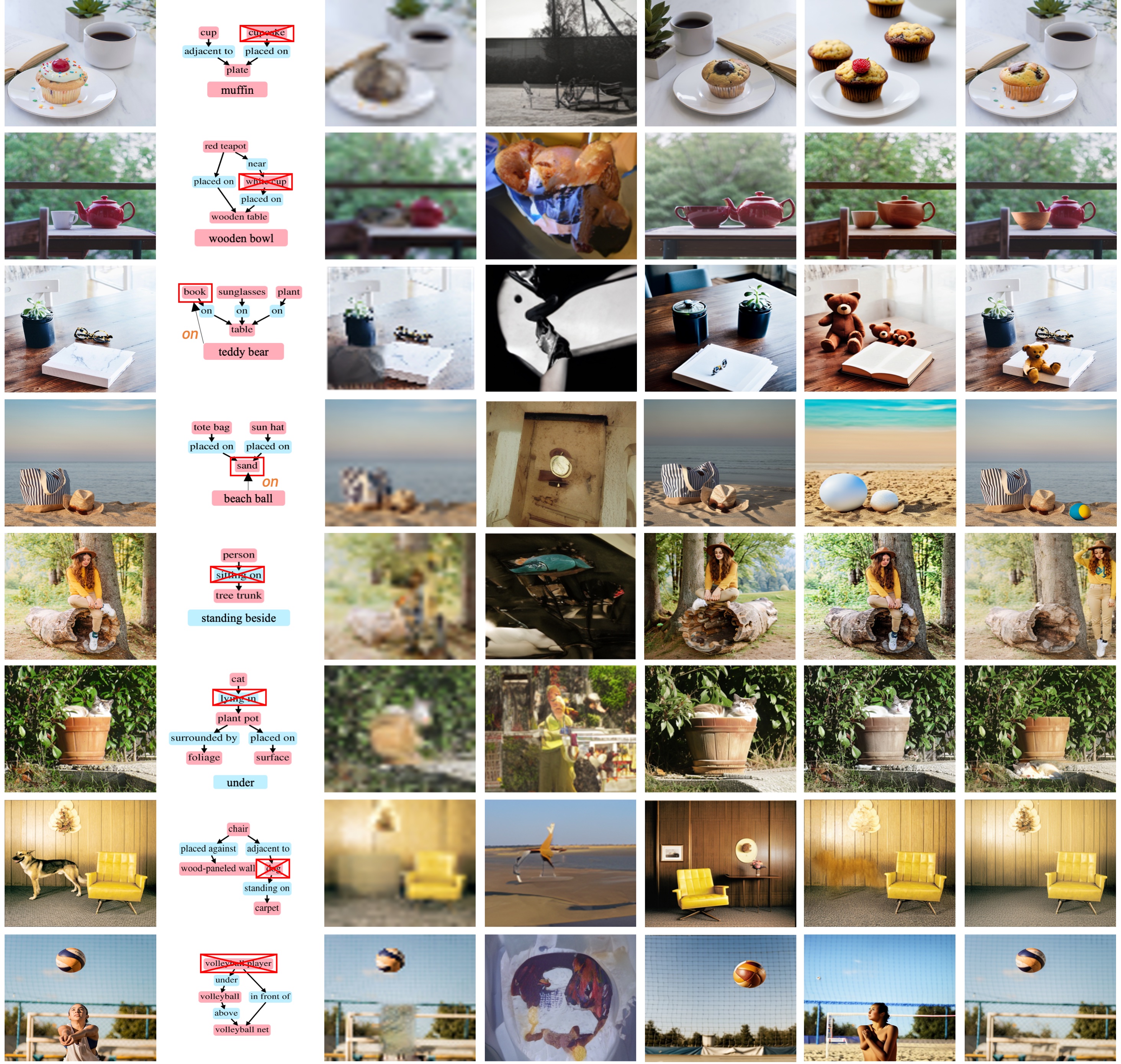}
    \parbox[t]{1.0\linewidth}{\relax
                           \hspace{28px} (a) \hspace{60px} 
                           (b) \hspace{58px} (c) \hspace{60px} (d) \hspace{60px} (e) \hspace{60px} (f) \hspace{60px} (g)}
    \caption{Qualitative comparison with other baseline methods. From left to right: (a) Input images; (b) Scene graphs and user edits; (c) SIMSG~\cite{dhamo2020semantic}; (d) SGDiff~\cite{yang2022diffusion}; (e) Break-a-scene~\cite{avrahami2023break}; (f) InstructPix2Pix~\cite{brooks_instructpix2pix_2023}; (g) Ours. Input images: the \textit{1\textsuperscript{st}}, \textit{4\textsuperscript{th}}, and \textit{6\textsuperscript{th}} rows are from \textcopyright iStockphoto; the \textit{2\textsuperscript{nd}}, \textit{3\textsuperscript{rd}}, \textit{5\textsuperscript{th}}, \textit{7\textsuperscript{th}}, and \textit{8\textsuperscript{th}} rows are from \textcopyright Unsplash.}
    \label{fig:comparison}
\end{figure*}

\subsubsection{Baselines}\label{sec:baseline}
In this comparison, we evaluate our method against four methods that can be adapted to support this task, including two scene-graph-based approaches and two state-of-the-art text-based image editing methods. For a fair comparison, we applied our constructed scene graph for all methods.

\cd{
We utilize two scene-graph-based approaches: SIMSG~\cite{dhamo2020semantic} and SGDiff~\cite{yang2022diffusion}, which are training-based methods leveraging GANs and diffusion models, respectively. To align with their training categories, we adapt our scene graph accordingly. Since SIMSG~\cite{dhamo2020semantic} relies on predefined bounding boxes for adding objects, we integrate LayoutTransformer to suggest placement regions, maintaining the original implementation for other functions.

For the diffusion-based methods, we select Break-A-Scene~\cite{avrahami2023break} and InstructP2P~\cite{brooks_instructpix2pix_2023}, converting our scene graph into prompts and instructions to suit their application scenarios. In Break-A-Scene, we learn distinct foreground tokens for each object and a <bg> token to represent the background. The scene graph is then transformed into text by prompting ChatGPT with 'make a short sentence based on the given scene graph, [in-context examples], [input scene graph].' For InstructPix2Pix, user modifications are naturally translated into specific instructions, such as 'add ... to ...' for addition, 'replace ... with ...' for replacement, 'remove ...' for deletion, and 'make ... [new relationship] ...' for modifying relationships.
}

\subsubsection{Evaluation Metrics} \label{sec:eval-metric}
As there is no ground truth for directly evaluating scene-graph-based image editing, we measure it from three aspects: element composition, relationship alignment, and image quality. Element composition and relationship alignment evaluate whether the objects and relationships in the edited image match the input image and accurately reflect the changes in the scene graph, respectively. Meanwhile, image quality measures the overall quality of the edited image. We implement metrics from these three aspects through both user study questions and GPT evaluations.



\subsection{Quantitative Evaluation.}
\ccd{We apply four types of edits (adding, replacing, removing objects, and modifying relationships) to 30 images randomly collected from the internet, each containing 2-6 common foreground objects, including diverse scenes of people, animals, vehicles, and everyday objects.} In total, our evaluation comprises 120 unique scene graph edits for each baseline and our method.

\subsubsection{User Study} \label{sec:user_study}
In this section, we conduct a user study to evaluate the effectiveness of our approach. To keep the user study manageable, we randomly select 10 edits from each of the four editing operations for human evaluation. For each question, we first show the original input image and the user's editing operation on the scene graph to the participants. Then, we display the five images edited by the four comparison methods and our method in a random order. Finally, participants are asked to select one option out of the five provided to answer three questions.

\begin{table}[t]
\centering
\caption{Human and GPT evaluation results from our user study are presented in the columns. The columns labeled EC, RA, and IQ represent the winning rates for element composition, relational alignment, and image quality, respectively, in the human user study. The columns labeled \(\tilde{EC}\), \(\tilde{RA}\), and \(\tilde{IQ}\) represent the scores for element composition, relational alignment, and image quality in the GPT4-V user study.}
\label{tab: userstudy}
\begin{tabular}{lcccccc}
\hline
\multicolumn{7}{c}{\textbf{Human and GPT Evaluation}} \\ 
\hline \noalign{\vskip 2pt}
& \textbf{EC} & \(\tilde{\textbf{EC}}\) & \textbf{RA} & \(\tilde{\textbf{RA}}\) & \textbf{IQ} & \(\tilde{\textbf{IQ}}\) \\
\hline
SIMSG & 0.01 & 0.76 & 0.01 & 0.66 & 0.01 & 0.52 \\
DiffSG & 0.00 & 0.19 & 0.00 & 0.09 & 0.01 & 0.37 \\
InstructP2P & 0.08 & 0.79 & 0.07 & 0.66 & 0.11 & 0.81 \\
Break-a-scene & 0.11 & 0.86 & 0.1 & 0.82 & 0.19 & 0.88 \\
Ours & \textbf{0.80} & \textbf{0.96} & \textbf{0.82} & \textbf{0.90} & \textbf{0.69} & \textbf{0.89} \\
\hline
\end{tabular}
\end{table}


\begin{itemize}
\item Element composition: Select the image that best preserves the composition integrity and visual identity of objects in the input image while reflecting the specified edits in the scene graph.
\item Relationship compliance: Select the image that best fulfills the specified relationship edits in the scene graph, while also maintaining other relationships as presented in the input image.
\item Image quality: Choose the image that demonstrates the highest overall quality and photorealism.
\end{itemize}

\zy{We collected a total of 26 questionnaires, with 14 male and 12 female participants of ages ranging from 18 to 45.} The results of the user study, as shown in Tab.~\ref{tab: userstudy}, indicate that our method outperforms the other methods in all three factors, achieving preferred rates of 0.80 in Element Composition, 0.82 in Relationship Compliance, and 0.69 in Image Quality.


\subsubsection{GPT-4V Evaluation} \label{sec:gpt-eval-intro}

Recent studies have demonstrated the potential of GPT-4V as a "generalist evaluator," showing great promise in conducting evaluations for various image generation tasks~\cite{zhang2023gpt}. In our experiment, we also utilize GPT-4V as a component of the evaluation process. Specifically, we design an evaluation task for each metric and provide the input-edited image pair, along with the task specification, to GPT-4V. The task specification comprises a task description, evaluation guidelines, and the target output format.

For each metric, instead of directly assigning a score, we require GPT-4V to review items in a checklist. For element composition, the checklist specifies which elements to add, remove, and preserve. GPT-4V is instructed to count the number of each element in the edited image before scoring. The score for each item ranges from 0 to 3. Deductions are made if the required elements are missing or over-represented, or if elements that should be removed are still present. Additionally, penalties are applied if the appearance of objects to be preserved varies from those in the original image. For relationship alignment, we again present a checklist by converting the scene graph into (subject, predicate, object) triples and instruct GPT-4V to review each relationship individually, determining if it is correct (score of 3) or missing (score of 0). Lastly, the checklist specifies four aspects of image quality: anomalies such as extra limbs, the texture and color of foreground objects, background lighting, and the overall realism of the scene. The score for each aspect ranges from 0 to 2, where 2 is the best and 0 is the worst. As we have a checklist for each of the three metrics, we average the scores of the checklist items to determine the final score for each metric.

As shown in Tab.~\ref{tab: userstudy}, our method outperforms other methods in all three factors, with scores of 0.96 for element composition, 0.90 for relationship alignment, and 0.89 for image quality. Additionally, we calculate the Pearson correlations between the winning rates from human evaluations and the scores derived from GPT-4V, which are 0.541, 0.557, and 0.669 for element composition, relational alignment, and image quality, respectively. The significantly high correlations with them affirm the validity of GPT-4V evaluation.

\subsection{Qualitative Comparison} \label{sec:qualitative}
Fig. \ref{sec:qualitative} presents qualitative comparisons between our method and four baseline approaches. Regarding SIMSG~\cite{dhamo2020semantic}, editing can be restricted to specific regions. However, due to the limited capacities of GANs, it produces images with the poorest quality, and the modified content in those regions is indiscernible. Furthermore, it introduces inconsistent squares that do not blend well with the background, as shown in rows 7 and 8. SGDiff~\cite{yang2022diffusion} is designed for image generation instead of image editing, so it generates images with completely altered content, as depicted in column (d). This model also struggles with generalizing to customized scene graphs, as the limited categories in its training dataset VG~\cite{krishna2017visual} and COCO~\cite{lin2014microsoft} constrain its ability to generate scenes with diverse objects and relationships. Break-A-Scene~\cite{avrahami2023break} encounters difficulties in composing complex scenes, such as failing to replace a cake with a muffin in row 1 or not adding a teddy bear in row 3. It also faces challenges in forming new relationships between objects, such as a person sitting on a tree instead of beside it in row 5 or a cat being on top of the pot instead of underneath it in row 6. Notably, it also alters parts of the background that should remain unchanged, such as repositioning the coffee cup and potted plant in row 1 or modifying the lighting and color of the sky in row 8. On the other hand, InstructPix2Pix~\cite{brooks_instructpix2pix_2023} exhibits inadequate localized editing capabilities. For example, in row 1, all objects on a table are transformed into muffins, and in row 4, beach objects are replaced with balls instead of adding balls. Its limited grounding is further demonstrated in the last row, where it removes volleyballs instead of volleyball players. Moreover, it fails to adjust object relationships, as shown in the editing examples in rows 5 and 6. In contrast, our method successfully conducts four types of edits, best preserves the visual identity of the input image, and generates the highest quality results.

\subsection{Ablation Study} \label{sec:ablation}

\subsubsection{Concept learning} \label{sec:ablation_inv}
\begin{figure}[htbp]
	\centerline{\includegraphics[width=1.0\linewidth]{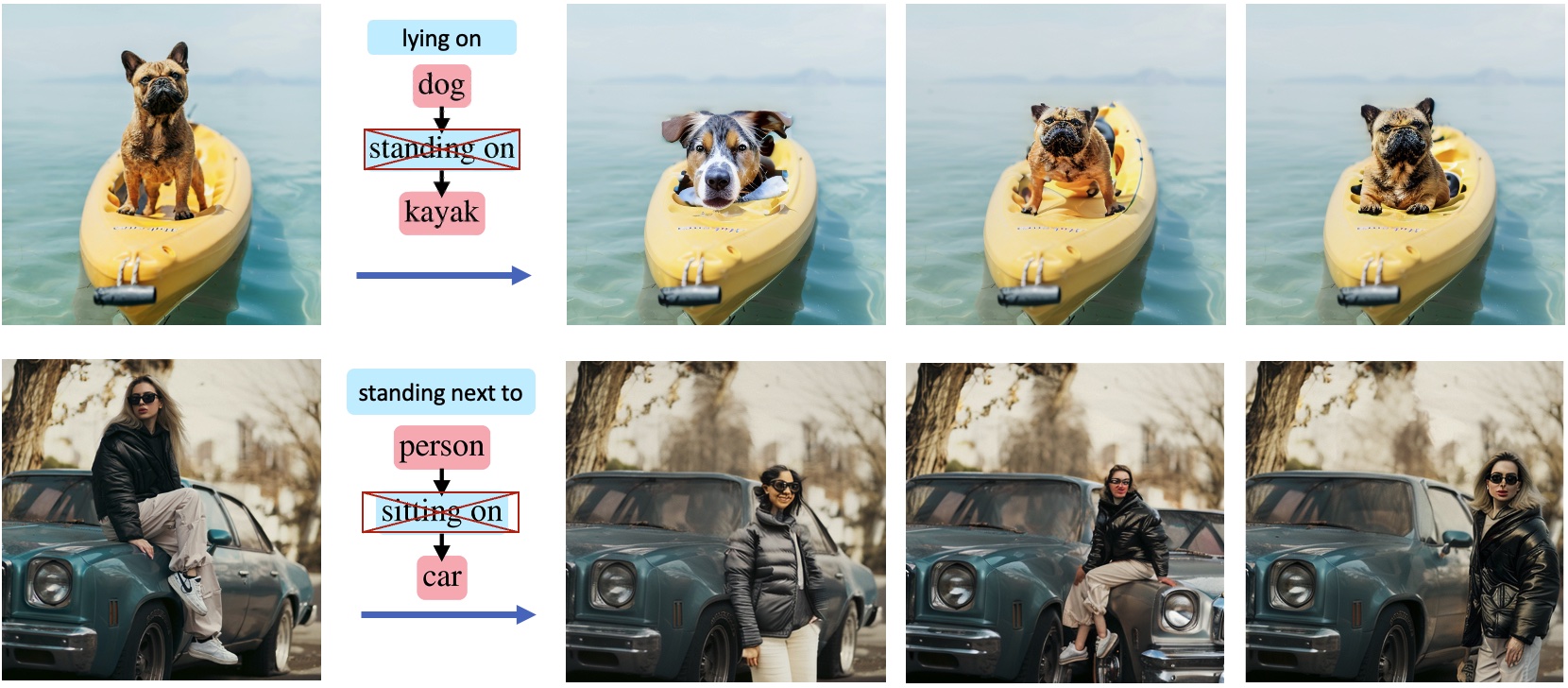}}
         \parbox[t]{1.0\linewidth}{\relax
                           \hspace{20px} (a) \hspace{28px} 
                           (b) \hspace{28px} 
                           (c) \hspace{38px} (d) \hspace{38px} (e)}
	\caption{Ablation study on different concept learning methods. From left to right: (a) Input images; (b) Simplified scene graphs and user edits; (c) Detailed prompt without textual inversion; (d) Textual inversion without detailed prompt; (e) Ours, which uses a detailed description plus textual inversion. The input images are from \textcopyright Unsplash.}
 \label{fig:ablation_inv1}
\end{figure}

To evaluate the effectiveness of our concept learning method with both detailed prompts and textual inversion, we compared it with two baselines, as shown in Fig. \ref{fig:ablation_inv1}: the first baseline only uses detailed prompts, as in the Idea2img method \cite{yang2023idea2img}, and the second baseline only employs textual inversion, as in the 'Break-a-Scene' method \cite{avrahami2023break}. Using a detailed prompt without textual inversion does not maintain object identity, as evidenced by the altered appearances of a 'dog' and a person in Fig. \ref{fig:ablation_inv1} (c). Meanwhile, relying solely on textual inversion leads to overfitting to fixed poses, resulting in incorrect poses of the dog and person in the edited image, as shown in Fig. \ref{fig:ablation_inv1} (d). Additionally, without the assistance of a detailed textual description, reconstructing all the details through textual inversion becomes difficult, leading to poor facial details of the person in Fig. \ref{fig:ablation_inv1} (d). In contrast, our method better preserves the visual properties and maintains editability, such as pose changes, as demonstrated in column (e).

\begin{figure}[htbp]
	\centerline{\includegraphics[width=1.0\linewidth]{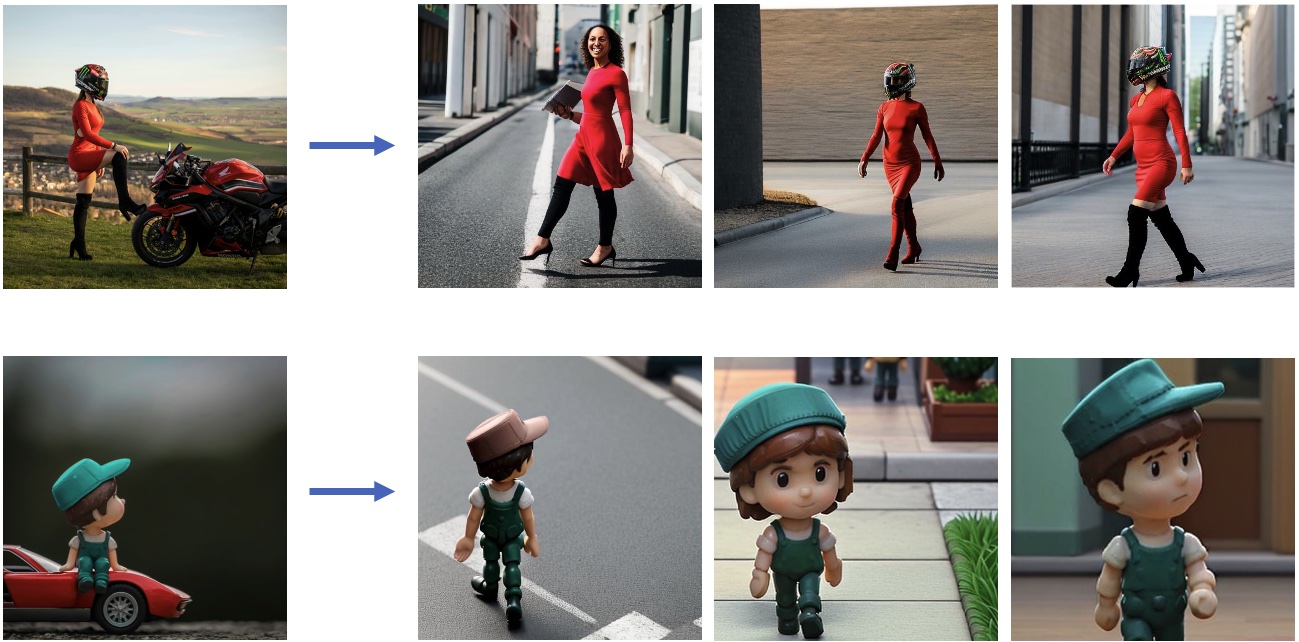}}
        \parbox[t]{1.0\linewidth}{\relax
                           \hspace{20px} (a) \hspace{65px} 
                           (b) \hspace{38px} (c) \hspace{38px} (d)}
	\caption{Ablation study on concept learning with prompts of varying levels of detail. From left to right: (a) Input images; (b) No detail: "A photo of a <opt>". (c) Medium detail: row 1, "<opt>...She is wearing a motorcycle helmet, a tight red dress."; row 2, "<opt>...The character has brown hair." (d) Full detail: row 1, "<opt>...She is wearing a motorcycle helmet, a tight red dress and black knee-high boots."; row 2: "<opt>...The character has brown hair and is wearing a teal cap." The input images are from \textcopyright Unsplash.}
 \label{fig:ablation_inv2}
 \end{figure}

Next, we examine the effect of varying levels of detail in the prompt, as shown in Fig. \ref{fig:ablation_inv2}. The images are generated by a customized diffusion model after concept learning with different prompts. Taking the first row as an example, when using "a photo of a <asset0>" without any detailed description, the female loses the helmet and black boots. With a medium level of detail in the description, the female retains the helmet, but the boots change. However, with our full training prompt, it best preserves the visual features. This suggests that adopting prompts with a higher level of detail allows the model to better capture the visual features.

\subsubsection{Graph-to-layout generation} \label{sec:ablation_position}
\begin{figure}[htbp]
	\centerline{\includegraphics[width=1.0\linewidth]{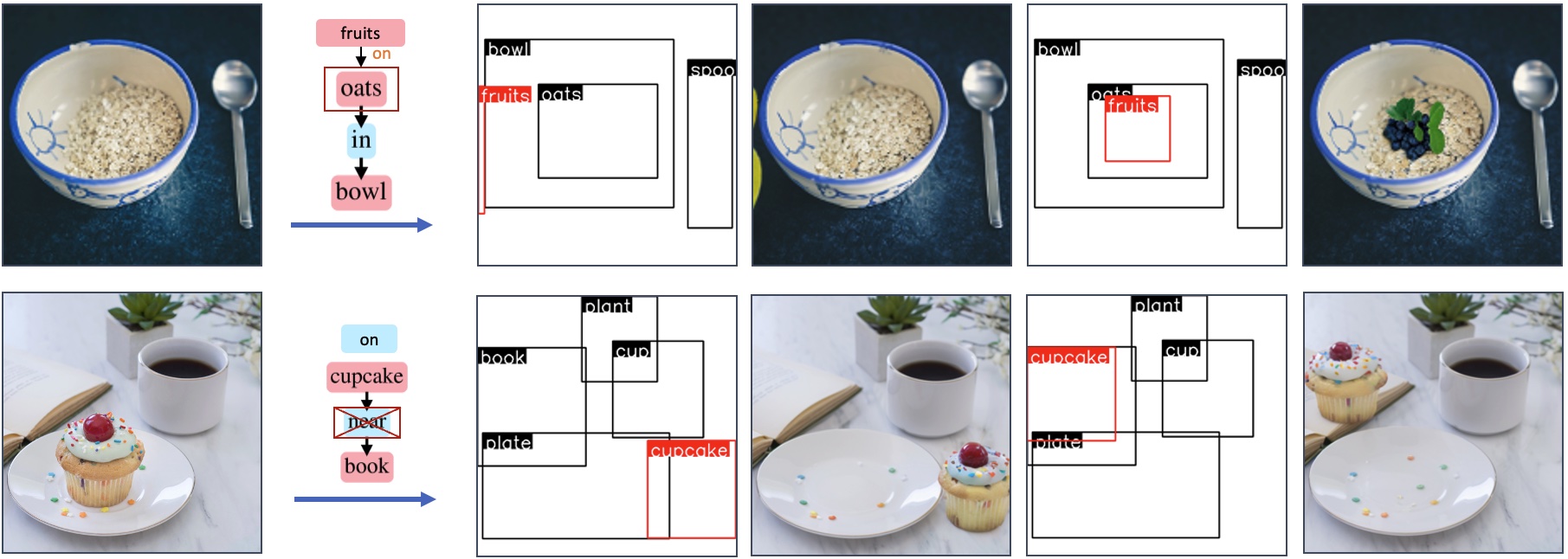}}
         \parbox[t]{1.0\linewidth}{\relax
                           \hspace{15px} (a) \hspace{20px} 
                           (b) \hspace{24px} (c) \hspace{28px} (d)\hspace{30px} (e)\hspace{30px} (f)}
	\caption{Ablation study on scene graph-to-layout generation. From left to right: (a) Input images; (b) Simplified scene graphs and user edits; (c) Layouts generated by Layout-Transformer (LT); (d) Images generated from LT layout; (e) Layouts generated by LLM; (f) Images generated from LLM layout (ours). Input images: the \textit{1\textsuperscript{st}} row is from \textcopyright Unsplash; the \textit{2\textsuperscript{nd}} row is from \textcopyright iStockphoto.}
 \label{fig:ablation_pos}
\end{figure}

After the user's edit on the scene graph, our LLM-driven editing controller predicts the bounding box to be edited. To evaluate its effectiveness, we compare it with a layout generation method, LayoutTransformer~\cite{gupta2021layouttransformer}, which generates the next layout element given the existing ones. To adapt LayoutTransformer for this task, we first map objects in the modified scene graph to its recognizable COCO categories and then transform the nodes in the scene graph with their bounding boxes into a layout elements sequence. For the objects to be added or altered, we obtain their positions from the output of LayoutTransformer. The results (Fig.~\ref{fig:ablation_pos}) indicate that the layouts predicted by this method have inaccuracies in positioning. For instance, in row 1, the fruits appear unnaturally thin and are positioned next to the bowl, and the cupcake is located in the lower right corner of the images, both of which violate the 'on' relationship. These inaccuracies stem from the model's sole reliance on the layout without considering the specific relational context between elements. In contrast, our method leverages LLM to explicitly comprehend both the layout and the inter-object relationships, as detailed in ~\ref{sec:planning}, enabling it to propose object placements more accurately.

\subsubsection{Object Removal} \label{sec:ablation_remove}
\begin{figure}[htbp]
	\centerline{\includegraphics[width=1.0\linewidth]{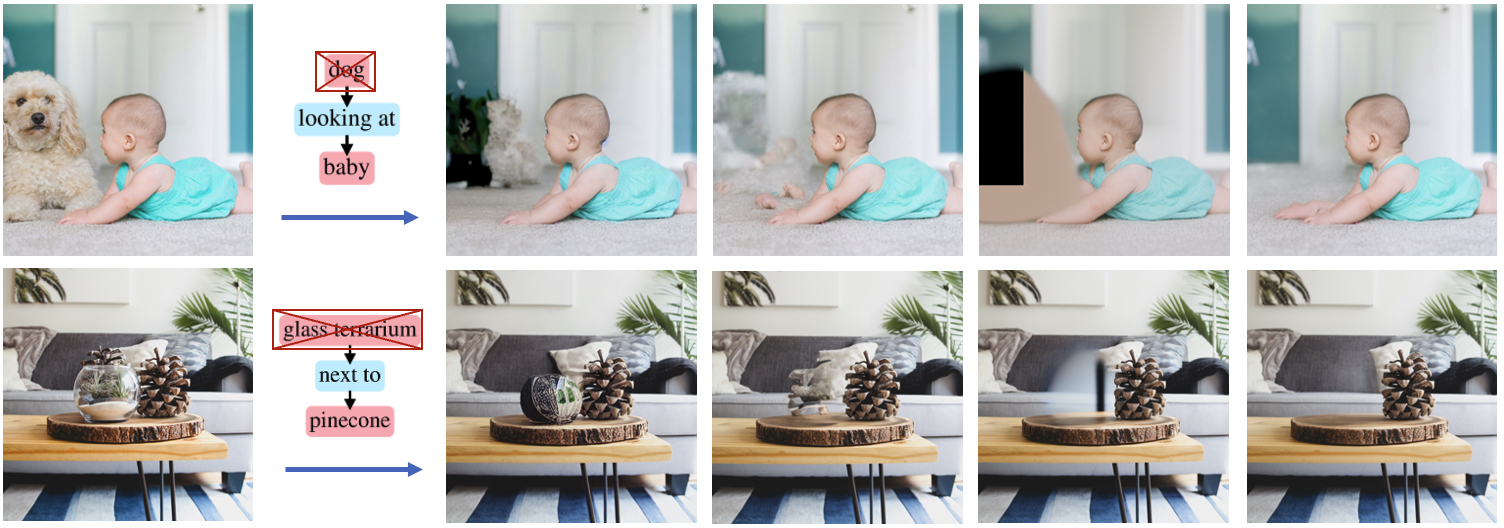}}
         \parbox[t]{1.0\linewidth}{\relax
                                   \hspace{12px} (a) \hspace{24px} 
                                   (b) \hspace{22px} (c) \hspace{28px} (d) \hspace{28px} (e) \hspace{30px} (f)}
	\caption{Ablation study on the object removal module. From left to right: (a) Input images; (b) Simplified scene graphs and user edits; (c) SD inpainting; (d) Without non-object prompt; (e) Without attention guidance; (f) Ours. The input images are from \textcopyright Unsplash.}
	\label{fig:ablation_remove}
\end{figure}
As described in Sec \ref{sec:removal}, our method applies a non-object prompt and attention modulation to directly remove objects using a Text2Image model. We first compared this approach with an alternative method using the Stable Diffusion inpainting model with a non-object prompt to remove objects directly. As shown in Fig. \ref{fig:ablation_remove} (c), it generates new elements in the target erased region instead of filling it with the background. This occurs because the training of the SD inpainting model does not distinguish between foreground and background, therefore it has no preference for generating background content. Next, we evaluated the effects of omitting the non-object prompt or attention modulation in our method. When using a null text prompt "" instead of the non-object prompt, there is an afterimage of the removed object as shown in \ref{fig:ablation_remove} (d). This happens because both the DDIM inversion and sampling use the same null text prompt, allowing information to leak through the cross-attention operation. Without attention guidance, although the object is removed, there is no consistency as there is no enforcement of the interaction between the newly generated region and the existing background. In contrast, our method can successfully erase the object and produce a clean and consistent background.

\subsubsection{Object Insertion}\label{sec:ablation_insertion}

\begin{figure}[htbp]
	\centerline{\includegraphics[width=1.0\linewidth]{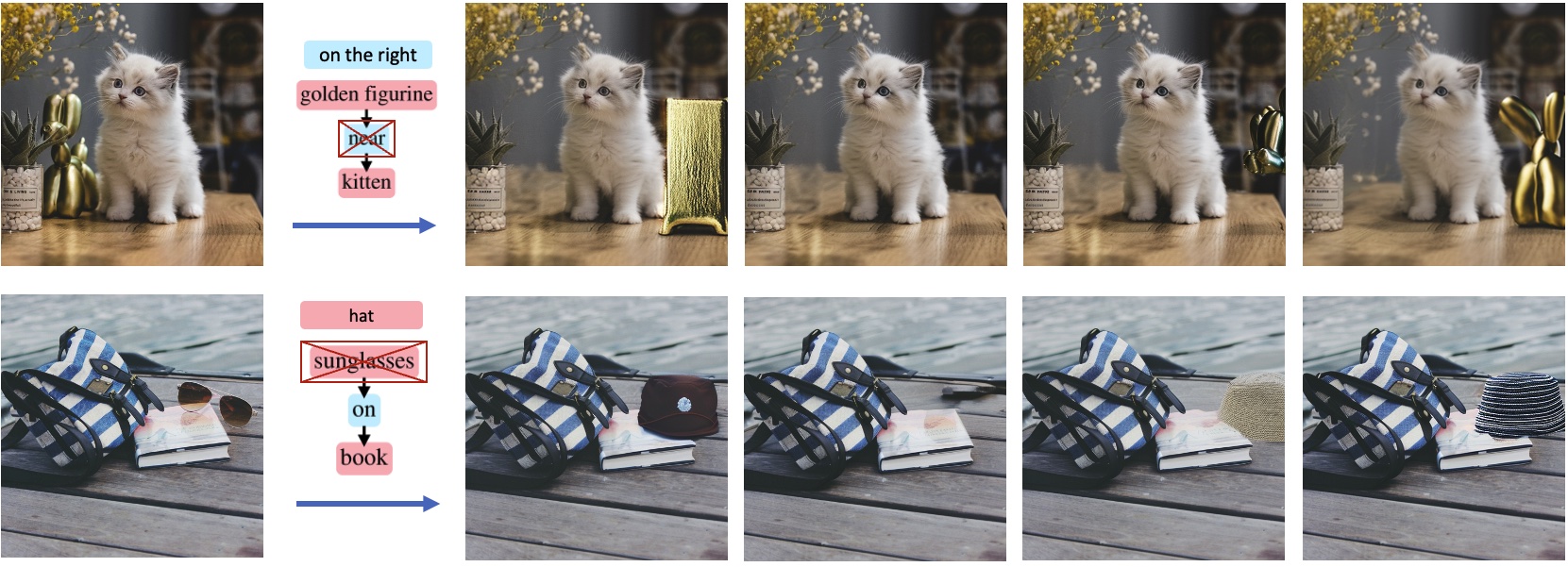}}
     \parbox[t]{1.0\linewidth}{\relax
                                   \hspace{12px} (a) \hspace{24px} 
                                   (b) \hspace{22px} (c) \hspace{28px} (d) \hspace{28px} (e) \hspace{30px} (f)}
	\caption{Ablation study on object insertion within a bounding box. From left to right: (a) Input images; (b) Simplified scene graphs and user edits; (c) ControlNet; (d) Blended Diffusion; (e) Dense-Diffusion; (f) Ours. The input images are from \textcopyright Unsplash.}
	\label{fig:ablation_layout}
\end{figure}

To verify the effectiveness of our attention modulation method for inserting the objects within the bounding box, we compared our approach with three other methods: ControlNet~\cite{zhang_adding_2023}, BLD~\cite{avrahami2023blended}, and DenseDiffusion~\cite{kim2023dense}. We convert our bounding \cd{boxes} into a semantic map as the layout condition for ControlNet and into binary masks for Blended Latent Diffusion (BLD) and DenseDiffusion. For ControlNet and DenseDiffusion, we use the same process as our method: pre-generating the objects before blending them with the background. For BLD, we use it to inpaint the object in the masked region. For the training-based method ControlNet, the generated objects tend to fill the entire bounding-box area, leading to unrealistic results due to conflicts with the common shape of the objects, such as the `golden figurine' in row 1 and `hat' in row 2 of Fig. \ref{fig:ablation_layout}. Applying BLD often results in generating the background instead of the target object because the attention values are not concentrated inside the small target region, leading to generation failures. Conversely, DenseDiffusion tends to generate objects larger than the target region, resulting in incomplete outcomes, such as the golden figurine and hat shown in \ref{fig:ablation_layout}. In contrast, our method applies additional regional constraints along with attention modulation, producing the best results in (f).


\begin{figure}[htbp]
	\centerline{\includegraphics[width=1.0\linewidth]{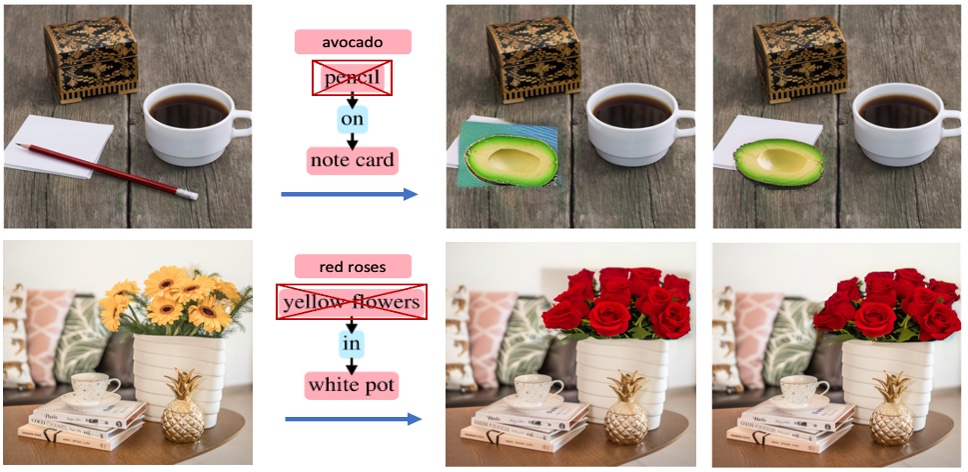}}
 \parbox[t]{1.0\linewidth}{\relax
                           \hspace{28px} (a) \hspace{40px} 
                           (b) \hspace{42px} (c) \hspace{50px} (d)}
	\caption{Ablation study on segmentation. From left to right: (a) Input images; (b) Simplified scene graphs and user edits; (c) w/o segmentation; (d) Ours. Input images: the \textit{1\textsuperscript{st}} row is from \textcopyright iStockphoto; the \textit{2\textsuperscript{nd}} row is from \textcopyright Unsplash.}
	\label{fig:ablation_seg}
\end{figure}
For blending inserted object with the \cd{background, we} segment the mask of the object region during the sampling stage to ensure the background \cd{consistency}. As shown in the Fig. \ref{fig:ablation_seg}, there is a faint blue rectangular border surrounding the avocado and grey surrounding the rose without applying the segmentation. Our method solves the aforementioned problems and generates \cd{better results}.

\begin{figure}[htbp]
	\centerline{\includegraphics[width=1.0\linewidth]{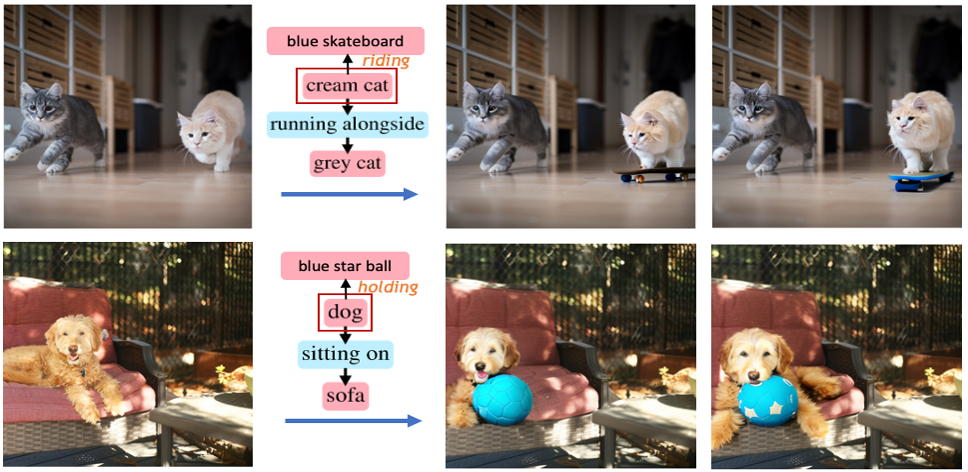}}
          \parbox[t]{1.0\linewidth}{\relax
                                   \hspace{28px} (a) \hspace{40px} 
                           (b) \hspace{42px} (c) \hspace{50px} (d)}
	\caption{Ablation study on multi-sampler. From left to right: (a) Input images; (b) Simplified scene graphs and user edits; (c) w/o multi-sampler; (d) Ours. Input images: the \textit{1\textsuperscript{st}} row is from \textcopyright iStockphoto; the \textit{2\textsuperscript{nd}} row is from \textcopyright Unsplash.}
	\label{fig:ablation_mult}
\end{figure}
We also apply the multi-sampler to minimize attribute leakage across instances when the insertion involves multiple objects. As shown in Fig. \ref{fig:ablation_mult}, without the multi-sampler, the blue skateboard loses the blue feature, and the blue star playball loses the star feature. The results in the rightmost column confirm that our multi-sampler minimizes attribute leakage.

\begin{table*}[ht]
\centering
\caption{Evaluation on the PIE-Bench Dataset~\cite{ju2024pnp}. Structure distance~\cite{tumanyan2023plug}, PSNR, LPIPS~\cite{zhang2018unreasonable}, MSE, SSIM~\cite{wang2004image}, and CLIP Similarity~\cite{wu2021godiva} are sourced from this benchmark. The columns labeled \(\tilde{EC}\), \(\tilde{RA}\), and \(\tilde{IQ}\) represent the scores for element composition, relational alignment, and image quality in our GPT-4V evaluation.
}
\resizebox{\textwidth}{!}{ 
\begin{tabular}{lcccccccccc}
\toprule
\textbf{Method} & \textbf{Structure} & \multicolumn{4}{c}{\textbf{Background Preservation}} & \textbf{CLIP Similarity} & \multicolumn{3}{c}{\textbf{GPT Evaluation}}\\
\cmidrule(r){2-2} \cmidrule(r){3-6} \cmidrule(r){7-7} \cmidrule(r){8-10}
& \textbf{Distance $\downarrow$} & \textbf{PSNR $\uparrow$} & \textbf{LPIPS $\downarrow$} & \textbf{MSE $\downarrow$} & \textbf{SSIM  $\uparrow$} & \textbf{Whole $\uparrow$} & \(\tilde{\textbf{EC}}\) \textbf{$\uparrow$} &  \(\tilde{\textbf{RA}}\) \textbf{$\uparrow$} & \(\tilde{\textbf{IQ}}\) \textbf{$\uparrow$}\\
\midrule
SIMSG & 0.08 & 19.48 & 0.40 & 0.01 & 0.70 & 20.40 & 0.71 & 0.53 & 0.41 \\
DiffSG & 0.17 & 9.35 & 0.55 & 0.14 & 0.41 & 12.45 & 0.26 & 0.13 & 0.31 \\
InstructP2P & 0.09 & 17.56 & 0.24 & 0.03 & 0.67 & 21.09 & 0.74 & 0.59 & 0.72 \\
Break-a-scene & 0.14 & 10.24 & 0.41 & 0.10 & 0.43 & 23.13 & 0.82 & 0.70 & \textbf{0.84} \\
Ours & \textbf{0.06} & \textbf{22.45} & \textbf{0.10} & \textbf{0.01} & \textbf{0.79} & \textbf{24.19} & \textbf{0.88} & \textbf{0.75} & 0.81 \\
\bottomrule
\end{tabular}
}
\label{tab: pie-bench}
\end{table*}

\begin{table}[ht]
\centering
\caption{Evaluation on the EditVal Dataset~\cite{basu2023editval}. DINO~\cite{caron2021emerging} and OwL-ViT~\cite{minderer2022simple} scores are sourced from the dataset. The columns \(\tilde{EC}\), \(\tilde{RA}\), and \(\tilde{IQ}\) correspond to our GPT-4V evaluation metrics, representing element composition, relational alignment, and image quality, respectively.
}
\begin{tabular}{lcccccccccc}
\toprule
\textbf{Method} & \textbf{Accuracy} & \textbf{Fidelity} & \multicolumn{3}{c}{\textbf{GPT Evaluation}}\\
\cmidrule(r){2-2} \cmidrule(r){3-3} \cmidrule(r){4-6} 
& \textbf{OwL-ViT} & \textbf{DINO} & \(\tilde{\textbf{EC}}\) & \(\tilde{\textbf{RA}}\) & \(\tilde{\textbf{IQ}}\)\\
\midrule

SIMSG & 0.11 & 0.57 & 0.66 & 0.55 & 0.38 \\
DiffSG & 0.01 & 0.13 & 0.27 & 0.14 & 0.30  \\
InstructP2P & 0.36 & 0.72 & 0.70 & 0.60 & 0.74 \\
Break-a-scene & 0.22 & 0.74 & 0.76 & 0.67 & 0.76  \\
ours & \textbf{0.53} & \textbf{0.83} & \textbf{0.91} & \textbf{0.85} & \textbf{0.76} \\
\bottomrule
\end{tabular}
\label{tab: editval}
\end{table}

\subsection{\ccd{Evaluation on Image Editing Benchmark}}
\cd{In addition to evaluating our method with images collected from the Internet, we also assessed its performance on two \ccd{public} image editing benchmarks: EditVal~\cite{basu2023editval} and PIE-Bench~\cite{ju2024pnp}. EditVal~\cite{basu2023editval} includes non-spatial edits, such as "Object-Replacement," and spatial edits, like "Positional-Addition" and "Positional-Replacement," which can be used to evaluate our replace, add, and modify edge functions. For PIE-Bench~\cite{ju2024pnp}, the change and delete object edits were used to assess our scene-graph editing method. However, we found that the add object edits in PIE-Bench are more focused on editing object parts rather than modifying the content of the scene. Therefore, we did not include them in our evaluation.

We used all the automatic evaluation metrics from each benchmark to assess the corresponding edits, except for FID in EditVal~\cite{basu2023editval} and regional CLIP similarity in PIE-Bench~\cite{ju2024pnp}. FID is generally not used for small sample sizes due to the high variance in statistical estimation~\cite{kynkaanniemi2019improved}. The regional CLIP similarity metric evaluates the similarity between the edited region and the description of the entire image, which is more relevant to other tasks included in the benchmark, such as changing the style of an image. Additionally, we incorporated our own GPT-4V evaluation for reference. The results are presented in Tab.~\ref{tab: pie-bench} and Tab.~\ref{tab: editval}.

In the EditVal experiments, our method outperformed others in editing accuracy, as demonstrated by higher scores in Owl-ViT (0.53), Element Composition (0.91), and Relation Alignment (0.85). Additionally, our approach maintained high image fidelity, as reflected in the DINO score (0.83) and Image Quality (0.76). On the PIE-Bench dataset, our method excelled in structure preservation (structure distance metric: 0.06) and background preservation (PSNR: 22.45; LPIPS: 0.1; MSE: 0.01; SSIM: 0.79). Furthermore, it achieved superior results in image editing, with higher CLIP text-to-image similarity (24.19), Element Composition (0.88), and Relation Alignment (0.75). Despite these improvements, our method maintained image quality comparable to Break-a-Scene, which regenerates the entire image.}

\section{Applications}
\begin{figure*}[t]
    \centering
    \includegraphics[width=1.0\textwidth]{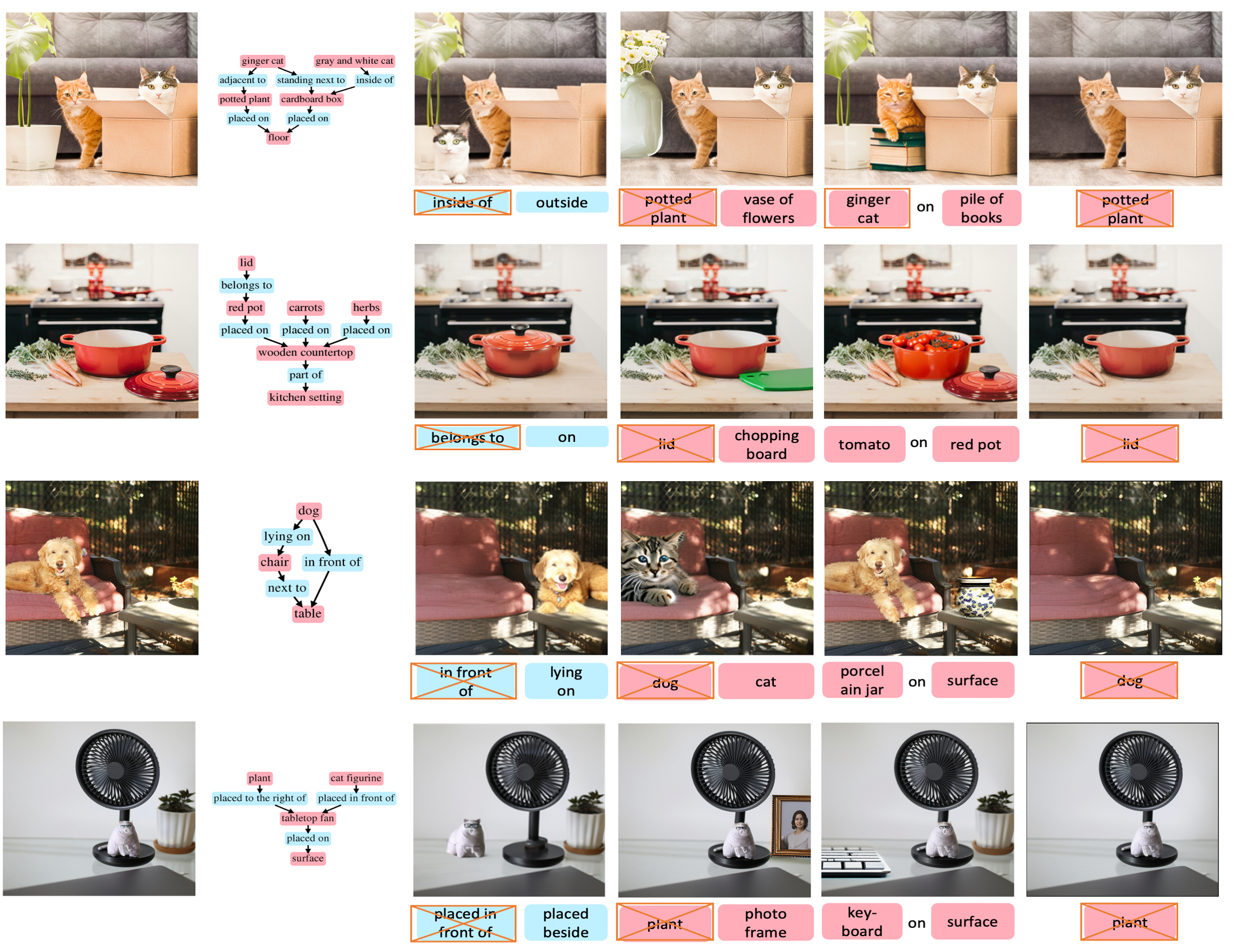}
    \parbox[t]{1.0\linewidth}{\relax
                                   \hspace{36px} (a) \hspace{65px} 
                                   (b) \hspace{75px} (c) \hspace{70px} (d) \hspace{70px} (e) \hspace{70px} (f)}
    \caption{Image manipulation with scene graph, including addition, replacement, removal, and relationship modification. From left to right: (a) Input images; (b) Scene graphs; (c) Relationship Modification; (d) Replacement; (e) Add; (f) Removal. Input images: the \textit{1\textsuperscript{st}} row is from \textcopyright iStockphoto; the \textit{2\textsuperscript{nd}}, \textit{3\textsuperscript{rd}}, \textit{4\textsuperscript{th}} rows are from \textcopyright Unsplash.}
    \label{fig:application}
\end{figure*}

Our method is capable of manipulating the scene graph to perform various kinds of image editing operations, including addition, replacement, removal, and relationship modification. It allows the user to explore the creative composition of a single input image through simple interactions with the scene graph. By leveraging LLM and text2image diffusion model pretrained on a huge dataset, our method is not limited to specific categories. Figs. \ref{fig:application} and \ref{fig:teaser} demonstrate some examples of our method's effectiveness. For instance, it can change the pose and position of an object while preserving its visual features, as demonstrated in row 4 of Fig. \ref{fig:application}. It can also alter objects in the images to a wide range of other objects, such as changing a lid to a chopping board in row 2 and adding a keyboard in row 3. Additionally, it can seamlessly remove objects from the scene, as illustrated in the last column. These examples highlight the flexibility of our method in handling various kinds of user editing operations.

\section{Discussions}
\begin{figure}[htbp]
    \centering
    \begin{subfigure}[b]{0.95\linewidth}
        \centering
        \includegraphics[width=\linewidth]{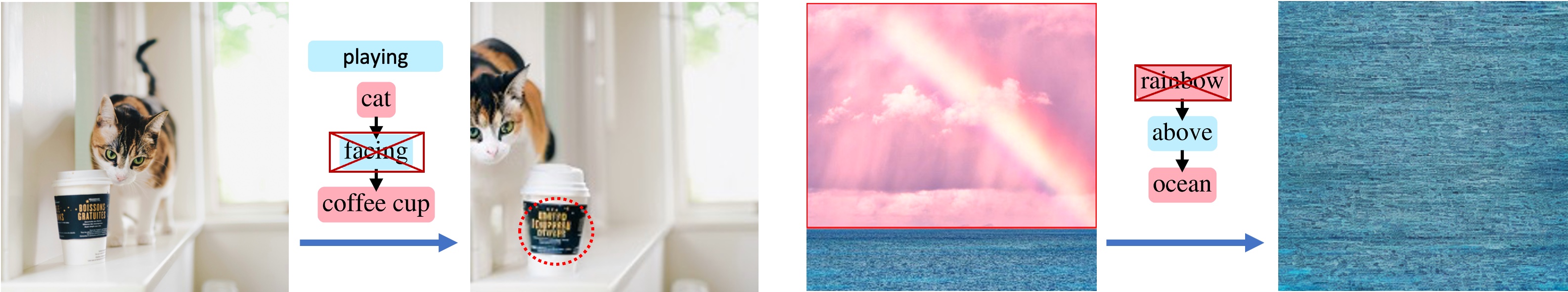}
        \caption{Examples of inaccuracies in concept learning and mask creation: The left image shows inadequate preservation of the object's appearance (e.g., the text on the cup contains artifacts), while the right image shows incorrect segmentation, erasing unintended regions (e.g., the entire sky is removed). The overlay mask highlights the rainbow segment. Input images: the left is from \textcopyright Unsplash; the right is from \textcopyright Pixabay.}
        \label{fig:limit_a}
    \end{subfigure}

    \vskip\baselineskip 
    \begin{subfigure}[b]{0.95\linewidth}
        \centering
        \includegraphics[width=\linewidth]{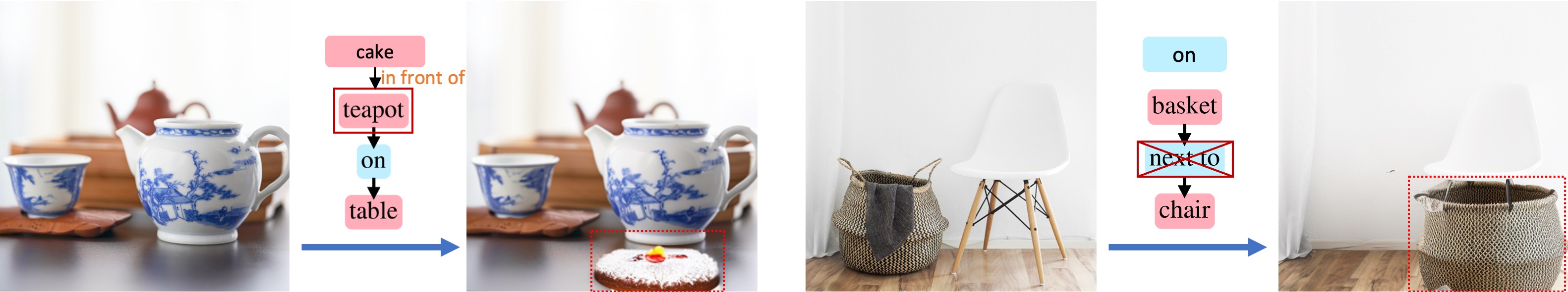}
        \caption{Examples of unsuitable bounding boxes proposed by the LLM, where the red dashed rectangle represents the predicted bounding box. Left shows incorrect size, right shows incorrect position. The input images are from \textcopyright Unsplash.}
        \label{fig:limit_b}
    \end{subfigure}

    \vskip\baselineskip 
    \begin{subfigure}[b]{0.95\linewidth}
        \centering
        \includegraphics[width=\linewidth]{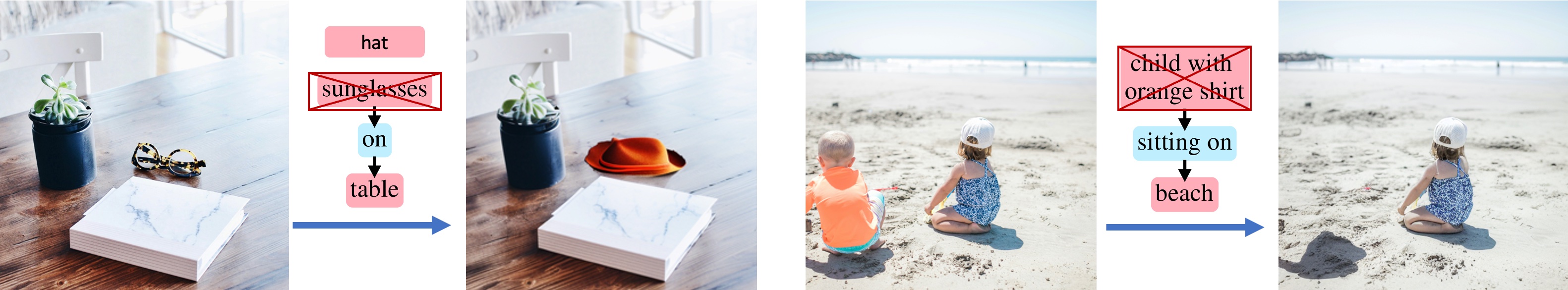}
        \caption{Examples of unrealistic lighting and shadows in image edits. The input images are from \textcopyright Unsplash.}
        \label{fig:limit_c}
    \end{subfigure}
    
    \caption{\ccd{Failure case}: (a) Inaccuracies in the scene parsing stage, where concept learning fails to fully replicate the object's appearance, and Grounded-SAM generates an incorrect mask. (b) The LLM fails to correctly infer the appropriate placement or size of the manipulated object based on the scene graph and bounding box information. (c) Lighting and shadows are not adjusted to match the modified objects, resulting in unrealistic visuals.}
    \label{fig:limitation}
\end{figure}

\textbf{Limitations} \cd{
Despite the advancements introduced by our framework, several limitations remain. First, the accuracy of our method is tied to the quality of scene parsing, as shown in the first row of Fig. \ref{fig:limitation}. When concept learning fails to replicate an object’s appearance precisely, the repositioned object may visually differ from the original image. Similarly, when Grounded-SAM generates incorrect object masks, subsequent editing operations are adversely affected, such as editing the wrong region. Second, the LLM may incorrectly infer the placement or size of manipulated objects, as shown in the second row of Fig. \ref{fig:limitation}. Since the LLM relies on bounding boxes and scene graphs for image context, details such as object parts (e.g., the seating surface of a chair) are not explicitly represented, making accurate spatial reasoning challenging. Another limitation is that the lighting and shadows generated by our method don't always match the modified objects, leading to unrealistic visuals, as shown in the third row of Fig. \ref{fig:limitation}. Lastly, our scene graph manipulation process is not real-time due to the need for fine-tuning the diffusion model.

These limitations may be addressed as base models like LLMs and diffusion models continue to advance. Future research could focus on expanding image editing applications and exploring iterative interactions between the LLM and the Text2Image model to enhance accuracy.
}

\textbf{Compatibility with Other LLMs} 
\cd{Our method is also compatible with other LLMs, including open-source ones. For scene graph construction, we achieved similar results using LLAVA-NEXT~\cite{liu2024llavanext}. For editing control, we applied the same edits to five images with both GPT-4~\cite{achiam2023gpt} and Mixtral-8x7B-Instruct-v0.1~\cite{jiang2024mixtral}. The average GPT evaluation scores from showed minimal difference (less than 5\%) between the two models, both outperforming Break-A-Scene, which had the highest performance among the baselines.}

\textbf{Compatibility with Other SD Version} 
\cd{Our method is adaptable to various versions of Stable Diffusion. To demonstrate this, we conducted 20 edits using SD-v1.5, with five edits for each type of manipulation. Additionally, we compared our results with InstructPix2Pix~\cite{brooks_instructpix2pix_2023}, which is fine-tuned from SD-v1.5 to showcase the superiority of our approach. All images will be included in the supplementary materials.}

\textbf{Societal Impact} 
\cd{Integrating scene graphs into image editing provides an intuitive interface for artists and designers to easily modify compositions and experiment with layouts. However, the ease with which images can be manipulated also raises potential ethical concerns, particularly regarding copyright and content ownership. The ability to reorganize and modify existing images could lead to unauthorized alterations or reproductions, potentially bypassing intellectual property rights, making it more difficult to ensure proper attribution and enforce ownership. To mitigate these risks, potential solutions such as embedding watermarks~\cite{luo2022leca} into generated images or employing forensic detection tools~\cite{corvi2023intriguing} to identify manipulated content can be considered.}

\section{Conclusions}
We present a novel scene graph-based image manipulation framework that integrates LLM and Text2image Generative model to enable sophisticated editing across a diverse range of descriptive terms. Our approach not only facilitates high-level, precise modifications of specific elements within an image but also supports the generation of creative and contextually coherent image alterations without the complexities of traditional image editing tools. Through the strategic use of LLM for scene parsing and editing control, our framework ensures that each edit is both flexible and context-aware. Moreover, we introduce an innovative "Attention-modulated Object Removal and Insertion" operation that leverages the capabilities of diffusion model for seamless object removal and spatially coherent object insertion. Additionally, our approach enhances the concept learning process with detailed prompts that ensure improved disentanglement and visual consistency in edited images. Our extensive experiments and perceptual studies demonstrate that our proposed method surpasses existing baseline methods in terms of composition elements, relationship alignment, and overall image quality.



\begin{acks}
We thank the anonymous reviewers for helping us to improve this paper. The work described in this paper was fully supported by a GRF grant from the Research Grants Council (RGC) of the Hong Kong Special Administrative Region, China [Project No.  CityU 11216122].
\end{acks}

\bibliographystyle{ACM-Reference-Format}
\bibliography{reference}
\appendix





\end{document}